# Herder Ants: Ant Colony Optimization with Aphids for Discrete Event-Triggered Dynamic Optimization Problems

**Authors:** Jonas Skackauskas*, Tatiana Kalganova*

College of Engineering, Design and Physical Sciences, Brunel University London*, United Kingdom

Abstract:

Currently available dynamic optimization strategies for Ant Colony Optimization (ACO) algorithm offer a trade-off of slower algorithm convergence or significant penalty to solution quality after each dynamic change occurs. This paper proposes a discrete dynamic optimization strategy called Ant Colony Optimization (ACO) with Aphids, modelled after a real-world symbiotic relationship between ants and aphids. ACO with Aphids strategy is designed to improve solution quality of discrete domain Dynamic Optimization Problems (DOPs) with event-triggered discrete dynamism. The proposed strategy aims to improve the inter-state convergence rate throughout the entire dynamic optimization. It does so by minimizing the fitness penalty and maximizing the convergence speed that occurs after the dynamic change. This strategy is tested against Full-Restart and Pheromone-Sharing strategies implemented on the same ACO core algorithm solving Dynamic Multidimensional Knapsack Problem (DMKP) benchmarks. ACO with Aphids has demonstrated superior performance over the Pheromone-Sharing strategy in every test on average gap reduced by 29.2%. Also, ACO with Aphids has outperformed the Full-Restart strategy for large datasets groups, and the overall average gap is reduced by 52.5%.

## 1 Introduction

Robust dynamic optimization approaches are of particular interest for real-world applications where problems at hand have a tendency to evolve over time and are not predictable accurately in advance. Often such problems must have some mechanisms to improve solutions in real-time as new data is obtained about disturbances or incremental accuracy improvements of the optimization data. Although the concept of dynamic optimization is not new, the research remains active to this day. Due to the real-world business interests in ever-increasing efficiency and growth, any incremental improvements in dynamic optimization are appreciated, according to a survey conducted by Mavrovouniotis et al. [1]. The swarm and evolutionary algorithms dominate the field of dynamic optimization. Genetic Algorithms (GA) [2] [3] [4], Particle Swarm Optimization (PSO) [5], Artificial Raindrop Algorithm (ARA) [6], Membrane Computing (MC) [7], and Ant Colony Optimization (ACO) [8], among many others [9] [10] [11] [12].

The increasing interest in the efficiency of Swarm Intelligence (SI) and Evolutionary Algorithms (EA) has led to the development of new optimization techniques. However, according to the No Free Lunch theorem, all of the proposed strategies show an equivalent performance when applied to all possible optimization problems [13]. The NFL theorem states that a general-purpose optimization algorithm cannot be regarded as a universally-best choice. As a result, the NFL encourages searching for more efficient methods and developing new optimization techniques and strategies for different fields.

## 1.1 Dynamic Optimization Problems (DOPs)

There are two major categories of the DOPs, shown in Figure 1, based on the problem search space: Continuous DOPs such as Multimodal Function Problems and Discrete DOPs such as Graph Problems. Both types of problems can have discrete or continuous dynamism. Continuous dynamism is a type of dynamism where the problem search space and objectives continuously transition or evolve throughout the optimisation process's entire duration. The transition is usually a predictable time-domain pattern that can be divided into arbitrarily small time steps. Problems with discrete dynamism are the problems where dynamic changes define independent problem states. Such state changes are usually caused by unpredictable external factors like event triggers.

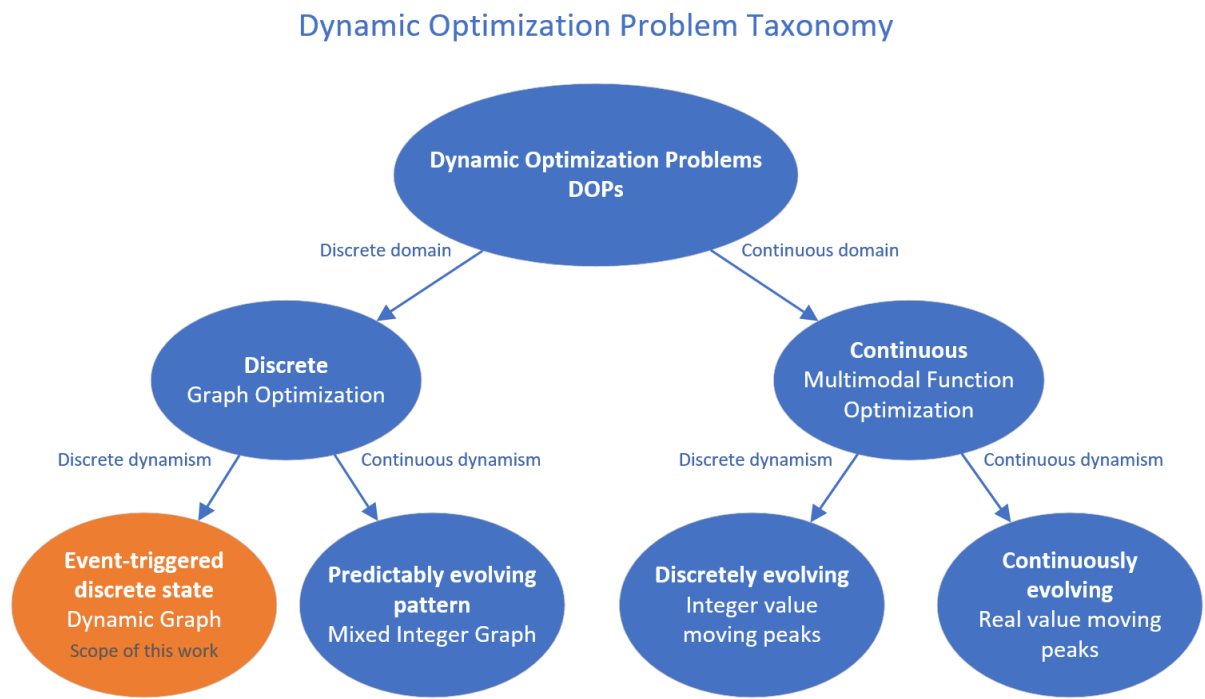

*Figure 1: Taxonomy diagram of Dynamic Optimization Problems (DOPs). The diagram emphasizes differences in problem input domain and dynamism. DOPs can be discrete or continuous domain problems and have discrete or continuous dynamism. This research focuses on the discrete domain with discrete dynamism type of problems.*

Continuous DOPs optimization is a widely researched topic with many published algorithms and improvements [1]. Continuous DOPs are modelled as function mappings from inputs to outputs with a dynamic time-domain parameter. Both the input and output variables are real numbers and subjected to constraints [14]. The time-domain component can be either a real value for a continuously evolving problem domain or an integer value for a discretely evolving problem domain. The Moving Peaks Benchmark (MPB) is a popular Continuous DOP type of benchmark with a wide selection of benchmark suites like IEEE CEC 2022 [15]. Such MPB suites can be used for both continuously evolving and discretely evolving problem setups.

This research will focus on solving the former type of Discrete Domain DOPs, in particular, Event-triggered Discrete DOPs where future problem instances are unpredictable, see Figure 2. These Discrete DOPs are modelled as the combination of discrete decisions and evaluated against fitness function. The dynamism of Discrete DOPs is defined as a series of static optimization problem instances in sequential order called "states" [16]. Each state has a slight variation of the search space, where the larger variation makes the problem more dynamic and more challenging to solve. Algorithms solving Discrete DOPs have been successfully applied to many theoretical and real-world optimization problems [17]. Genetic Algorithms (GA) are used to optimize traffic signal timing control [18] as well as to solve the Dynamic Multidimensional Knapsack Problem (DMKP) [19]. Particle Swarm Optimization (PSO) algorithms are used to optimize control parameters of a dynamic chemical process [20] as well as Dynamic Traveling Salesman Problem (DTSP) [21] [22]. Ant Colony Optimization (ACO) algorithms are applied to railway junctions scheduling problem [8] and DTSP [23].

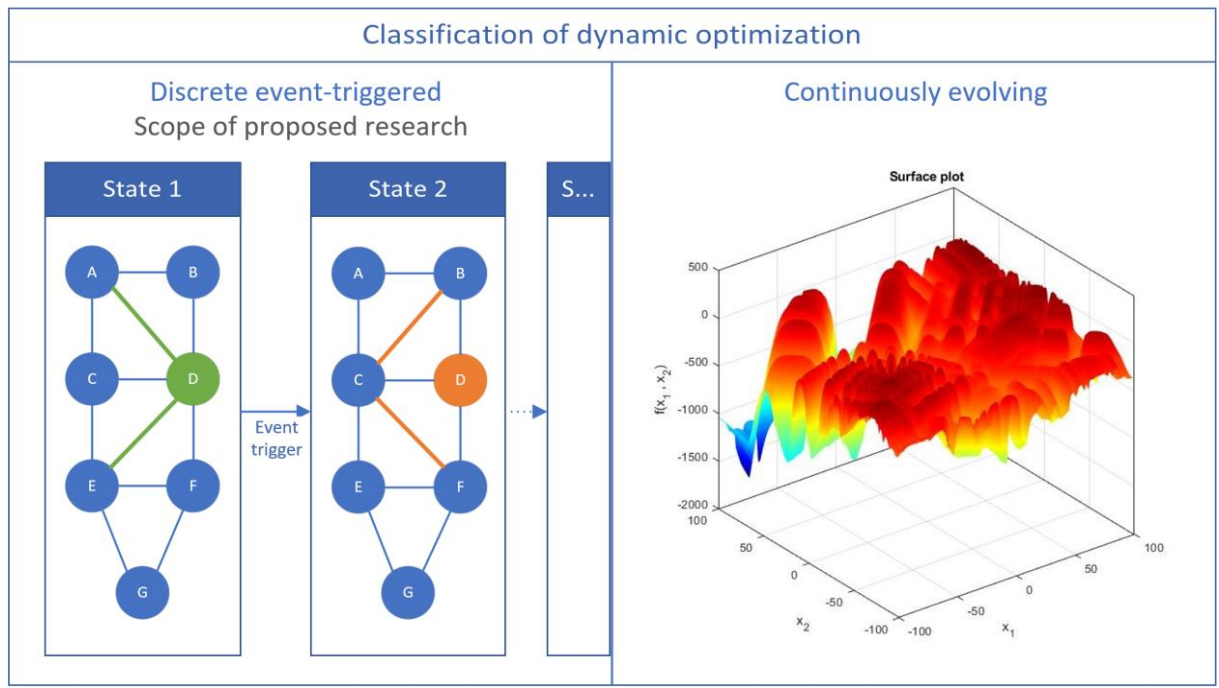

*Figure 2: Classification of dynamic optimization. On the left-hand side, Discrete dynamic optimization is displayed with a discrete optimization problem state change based on an event trigger. On the right-hand-side popular Continuous DOP, a moving peaks benchmark is visualized for two input dimensions from IEEE CEC 2022 benchmark set [15]. This research focuses on solving types of problems portrayed on the left-hand side.*

Up to recently, there was no existing Discrete event-triggered DOP benchmark to evaluate the efficiency of proposed algorithms, especially for a limited time per state of the dynamic optimization. This work uses recently proposed DMKP fully-defined benchmark datasets and evaluates the efficiency of several different ACO dynamic optimization strategies. Furthermore, this work proposes a new Aphids-based interstate transition technique to enhance overall performance on the DMKP benchmark.

## 1.2   Existing ACO methods to solve DOPs

Numerous studies have been conducted on the Ant Colony Optimization algorithm to solve problems of evolving nature [24] [25] [26]. In the classical Ant Colony Optimization algorithm, changing any aspect of the optimization problem leads to explored solutions becoming infeasible or suboptimal [27]. In nature, real ants constantly face environmental changes such as new food sources appearing or being removed and pathway blockage. However, real ants do not start the search completely from scratch after the environmental change. Hence artificial ants might not need to restart either. Usually, algorithms solving static optimization problems are finely tuned to quickly converge onto optimum solutions in the search static search space. However, such behaviour might not be desirable for dynamic optimization problems because fast and precise algorithms usually struggle to explore search space after the dynamic change. The challenge is to enable artificial ant colony to explore new changes in the problem space efficiently [28]. There are several explored methods of Ant Colony Optimization in the dynamic environment.

In dynamic optimization literature, research usually tackles one of the two types of dynamic optimization problems. The first type is the optimization problems executed in the time domain, which has predictable dynamism patterns. These patterns of problem dynamics are solved as an additional dimension of variable edge cost in single goal optimization. This type of optimization problem does not require a special dynamic optimization algorithm but requires modification for the optimization problem definition. Typical applications of such predictive optimization are in routing problems like Vehicle Routing Problem (VRP) [29] [30], electric grid energy management [31], or scheduling problems [32]. The second type of DOPs is unpredictable event-triggered DOPs. When dynamic environment changes cannot be predicted, the problem must be solved again. Two major strategies for the ACO algorithm have been researched to solve the unpredictable event-triggered DOPs, Full-Restart and Pheromone-Sharing strategies.

*Table 1: Related work summary of advantages and disadvantages of using ACO with Full-Restart and Pheromone-Sharing strategies.*

| Category | Publication | Advantages | Disadvantages |
|---|---|---|---|
| Full-Restart | [33] | Easy to implement. | Even smallest changes need to be reoptimized. |
| | [27] | Given enough time to optimize. The final results were better than Pheromone-Sharing. | Takes longer to converge to good solution. |
| | [26] | HACO algorithm utilizes daemon actions to improve results. | Runtime significantly increased for modest results improvement. |
| Pheromone-Sharing | [24] | Reduced computational burden in dynamic optimization compared to Genetic Algorithm. | The algorithm was slower than GA for static optimization. |
| | [25] | While sharing pheromone, using large ant population allowed to overcome local optimums. | Larger population requires more computational resource to converge. |
| | [27] | Shared pheromone allowed for compounding convergence. | Each state converged slightly slower than Full-Restart strategy. |
| | [34] | Shared population showed strong resilience towards dynamic changes. | Typical elitist population performs poorer than probabilistic population. |
| | [23] | P-ACO can perform well as long as population is diverse. | P-ACO is less stable than standard MMAS. |
| | [35] | Sharing population and pheromones showed increasingly better results with larger optimization problems. | When search space changes significantly, ants struggle to explore it adequately. |

### 1.2.1 Full-Restart strategy

The most common approach to solving dynamic optimization problems when changes are unpredictable is to restart the search entirely. When changes occur in search space, the whole optimization is started from the beginning, clearing out all information associated with the previous state. This approach is the simplest as it does not require explicit algorithm modification and relies on the algorithm to quickly converge to the new good solution [33]. The Full-Restart strategy does not share any information between optimization states, and each state is optimized independently. After the full restart algorithm usually converges in an identical pattern to the state before, as shown in Figure 3.

In the literature, Two ACO dynamic optimization strategies, Full-Restart and Pheromone-Sharing, have been compared by Angus & Hendtlass [27]. The conclusion was reached that for Dynamic Traveling Salesman Problem, the Pheromone-Sharing strategy has given a faster convergence rate to a good solution. Still, the Full-Restart strategy has allowed converging to a more optimal solution. Then, ACO algorithm Full-Restart strategy was used in a railway routing problem, and the solutions rebuild and deployed in real-time, allowing the algorithm to run for a limited amount of time, and deploy the best solution that the algorithm has found [36]. Furthermore, ACO Full-Restart strategy was investigated on real-time dynamic optimization problems, where optimal schedules and routes are already in use and partially completed. The dynamic changes alter the remaining part of the solution, which requires to be reoptimized for problems like VRP [26], and Job-shop Scheduling Problem (JSP) [37].

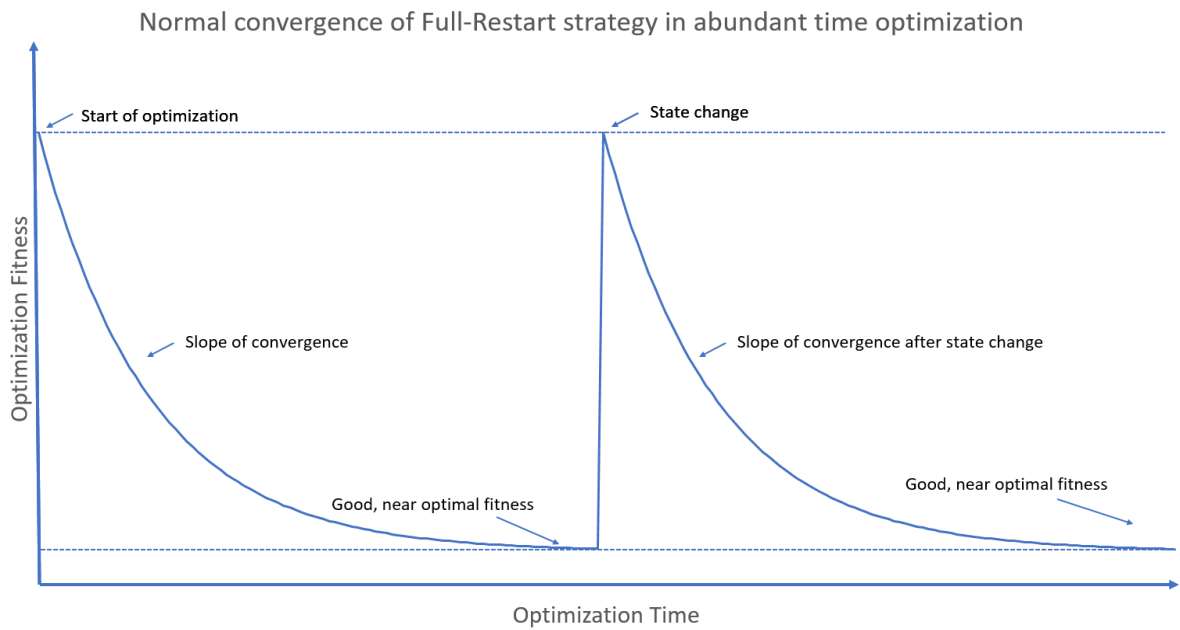

Figure 3: Normal Convergence of Full-Restart strategy in abundant time optimization. The chart displays a minimization problem's convergence for two dynamic states, where the algorithm has plenty of time to converge to a "good" solution in both instances. Both initial and following states show identical convergence patterns.

### 1.2.2 Pheromone-Sharing strategy

When unpredictable dynamic change occurs in the environment, the search space changes too. However, changes in the nodes and the edges of the new search space are usually small enough and can map to the old search space. Therefore, the artificial ants can reuse a large part or the whole pheromone matrix from the previous optimization state. In cases where dynamic changes are significant to the extent that some parts of search space do not map the pheromone matrix correctly, a heuristic fix can be applied to maximize the usefulness of the pheromone matrix from the previous state.

In literature, some ACO-based methods rebuild solutions on the existing pheromone matrix where new edges get applied to normalized pheromone level allowing ants to have a fair exploration in new search space [27]. Other development has tried several pheromone initialization strategies for new edges with the proposed Local random restart strategy, which initializes new edges with a random value, and the Local restart strategy, which initializes new edges with 0 pheromone value [25]. Approaches based on Population-ACO use the pheromone initialization process described by Guntsch & Middendorf [34], where an arbitrary number of elitist ant solutions create a pheromone matrix for every new iteration. Then Population-ACO based solutions can also be fixed using heuristic methods after the dynamic change, which give a good head start for the pheromone quality after the dynamic change [23] [35]. Generally, the Pheromone-Sharing strategy makes a trade-off, shown in Figure 4, to significantly reduce the fitness penalty after the dynamic change at the cost of reduced convergence speed due to the necessity to evaporate previous state dynamics.

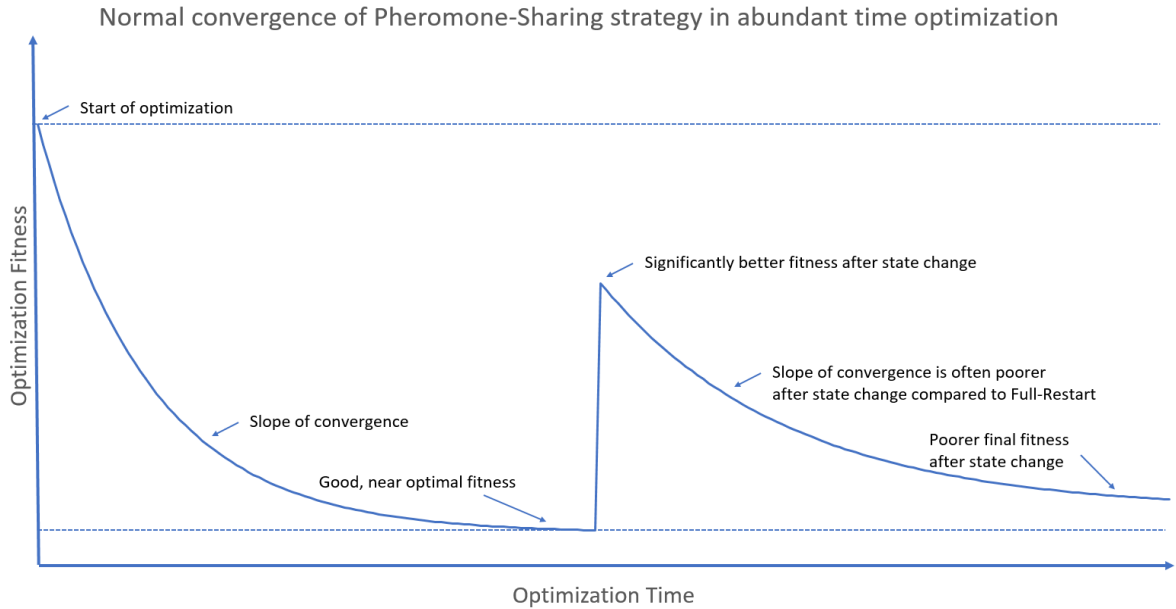

*Figure 4: Normal Convergence of Pheromone-Sharing strategy in abundant time optimization. The chart displays a minimization problem's convergence for two dynamic states, where the algorithm has plenty of time to converge to a "good" solution in both instances. The initial state converges normally, and the following state starts at a significantly better fitness level but shows poorer convergence. Then poorer convergence leads to poorer final fitness result.*

## 1.3  Need for discrete event-triggered dynamic optimization system

Both dynamic optimization strategies overviewed above have their specific advantages and disadvantages. The Pheromone-Sharing strategy significantly improves solution quality after the dynamic change because the pheromone matrix is reused for the new optimization state after the event triggers a dynamic change. When dynamic changes are triggered frequently, the Pheromone-Sharing strategy allows for gradual interstate convergence, which is better than Full-Restart strategy, see Figure 5. However, some portion of the pheromone from the previous state must evaporate gradually, slowing the discovery of new paths. Therefore, the convergence of the Pheromone-Sharing strategy is usually slightly worse than the convergence if the search is restarted. On the other hand, the Full-Restart strategy performs good quality optimization when dynamic changes are triggered very infrequently, and the algorithm has plenty of time for convergence. However, when dynamic changes occur frequently, the Full-Restart strategy carries no information and restarts the search completely with poor results, see Figure 5.

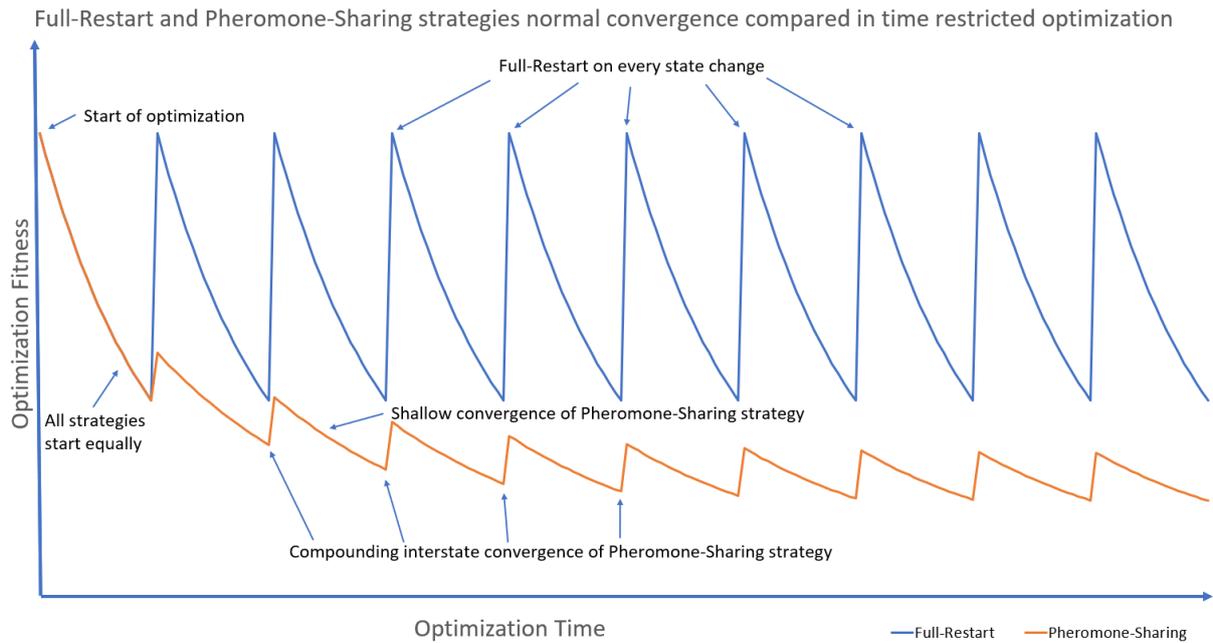

*Figure 5: Full-Restart and Pheromone-Sharing strategies normal convergence compared in time restricted optimization. The chart displays a minimization problem's convergence for ten dynamic states changing frequently. Full-Restart strategy shows great optimization convergence, but better restart of Pheromone-Sharing strategy allows for compounding interstate convergence.*

Ideally, a purpose-built dynamic optimization method could combine the strengths of the Full-Restart strategy's convergence speed and the Pheromone-Sharing strategy's good solution quality after the dynamic change. This work introduces a purpose-built dynamic optimization strategy for the ACO algorithm called "ACO with Aphids". The ACO with Aphids is a nature-inspired dynamic optimization strategy that builds robust optimization without drawbacks of poor convergence and poor restart after the dynamic change. In Figure 6, the ideal ACO with Aphids example is shown. In this example, after the dynamic change, fitness is equally good to the fitness of the Pheromone-Sharing strategy, but the convergence slope is as good as the Full-Restart strategy convergence.

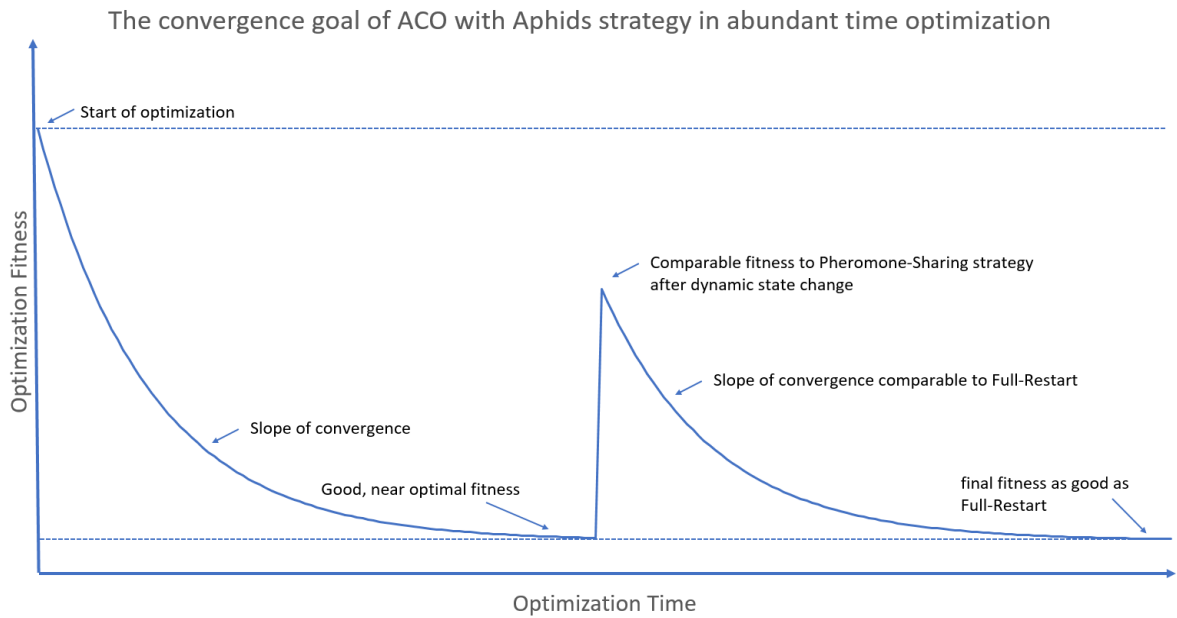

*Figure 6: The convergence goal of ACO with Aphids strategy in abundant time optimization. The chart displays a minimization problem's convergence for two dynamic states, where the algorithm has plenty of time to converge to a "good" solution in both instances. The initial state converges normally, and the following state starts at a significantly better fitness level similar to the Pheromone-Sharing strategy and shows equally good convergence to the Full-Restart strategy. Then a good restart fitness after the dynamic change and a good convergence lead to even better results than the final result of the Full-Restart strategy.*

A good dynamic optimization system should display the biggest solution quality improvements for frequently changing dynamic optimization. When the time is limited to perform the optimization of each state, at the start fitness usually does not converge to an acceptable level, but following states' optimization continues to improve the fitness further. ACO with Aphids strategy aims to intelligently reuse the information acquired during previous states' optimization such that after the dynamic change is triggered, the penalty to optimization fitness is minimal, and further convergence is unimpeded. This way, maximum interstate convergence equilibrium is possible, see Figure 7.

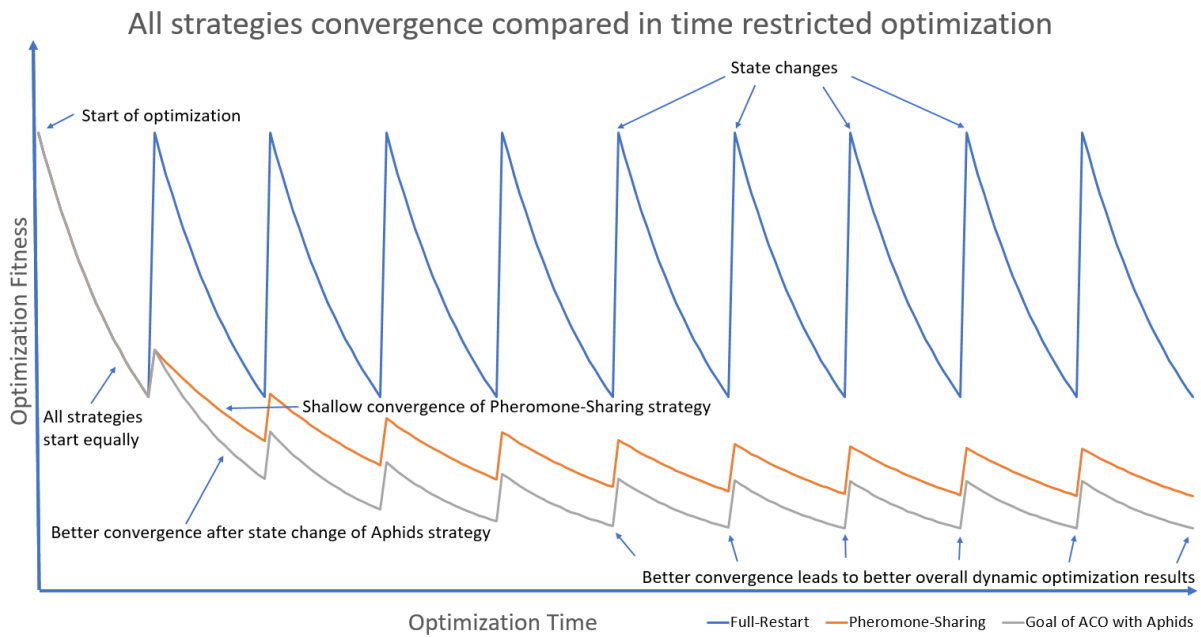

Figure 7: All strategies convergence compared in time-restricted optimization. The chart displays a minimization problem's convergence for ten dynamic states changing frequently. Aphids strategy combines great optimization convergence observed in Full-Restart strategy with low state change fitness penalty observed in Pheromone-Sharing strategy. The combination of these strengths allows for better interstate convergence.

## 1.4 Nature of Herder Ants

Some of the ant species, namely *Lasius niger* [38], care for and herd aphids. Aphids are tiny green bugs that feed on plants and produce honeydew as waste. The honeydew is a sugar-rich liquid that is very nutritious to ants and acts as an additional food source. The relationship between ants and aphids is symbiotic. While ants feed on the aphids' wasted honeydew, ants also protect aphids from their natural predators [39]. The pheromone laid down by ants has a behavioural effect on aphids. In the ants' pheromone presence, aphids move slower and produce more honeydew [40]. There was also observed that ants may prey on aphids based on aphid density, honeydew production, and how much other ants tended to aphids [41].

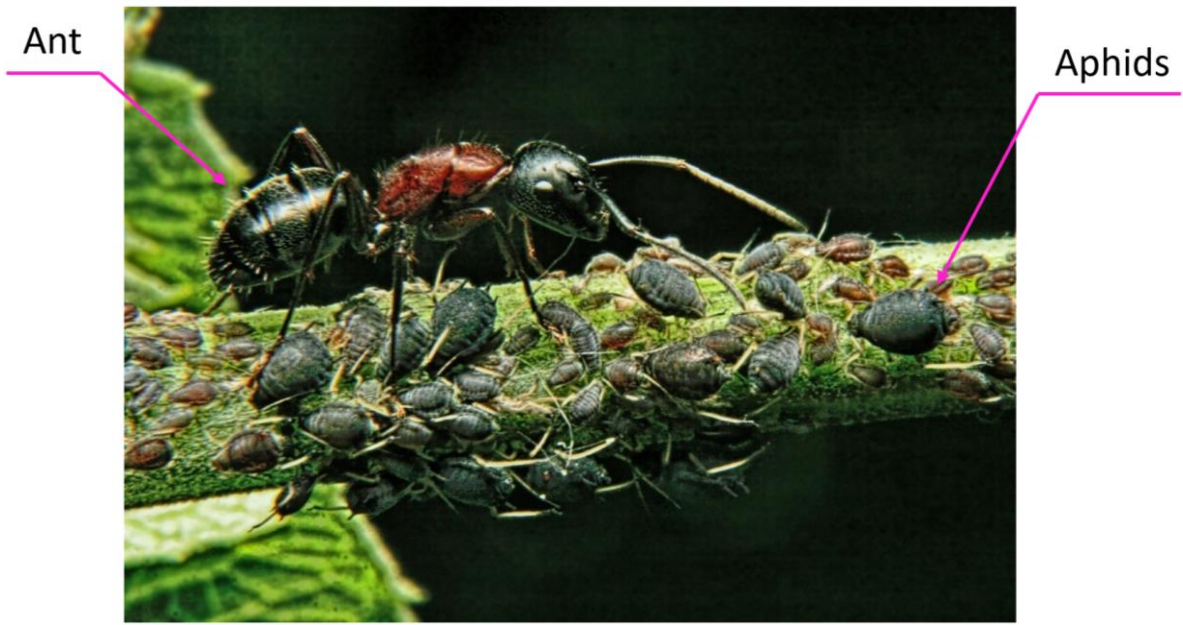

*Figure 8: Close-up image of an ant guarding its aphids [42].*

## 1.5 Use of aphids in other optimization algorithms

Aphid–Ant Mutualism (AAM) is a heterogeneous population-based algorithm that considers two types of individuals: ants and aphids [43]. In this algorithm, aphids perform a search in tandem with ants' search but with a different fitness function. Researchers have applied this algorithm to solve Multimodal Function Optimization benchmarks.

Cartesian Ant Programming (CAP) is a Cartesian Genetic Programming (CGP) algorithm that uses ants and aphids to optimize connections among function symbols [44]. In this algorithm, ants perform a search to find a valid set of connections and use the solution to deposit pheromone on each connection to attract more ants. In addition to pheromone, ants are also attracted to honeydew that is deposited on every node deposited by aphids.

Ant Colony Optimization with Cooperative Aphid (ACOCA) is a cooperative search method that combines the capabilities of an ant and aphid [45]. In this method, the ant and the aphid work together to give a solution. The information that the aphids provide is then treated as honey and the search solutions of the ant are influenced by the honey. This method was tested to solve Traveling Salesman Problem (TSP).

So far, a symbiotic relationship between ants and aphids has not been attempted to model to benefit the discrete dynamic optimization. Aphids are tiny animals that ants tend to by placing them on plants, protecting them from predators, and collecting their waste honeydew. When too many aphids exist, a portion of them is killed to maintain the optimal aphids population. In simplistic terms, aphid's honeydew production can represent the current state of the dynamic environment and use ants' behaviour of tending, relocating, and killing aphids to optimize for maximum food supply from the original ant's objective supplemented with honeydew supply from aphids.

## 1.6 Dynamic optimization research gap

So far, most theoretical research on dynamic optimization with comparable qualities has been done primarily on continuous domain dynamic optimization. The research on discrete domain dynamic optimization problems is either solving theoretical optimization problems by modifying datasets using stochastic methods or solving a real-world optimization problem. Optimization solutions of stochastically generated dynamic optimization datasets cannot be fairly compared because dataset results and optimal values will have some variance. Also, research on real-world optimization problems usually does not share the dataset used for the optimization. This work aims to comprehensively analyse two well-known ACO dynamic optimization strategies, Full-Restart and Pheromone-Sharing, using the DMKP benchmark. This work also proposes a new dynamic optimization strategy called ACO with Aphids to improve overall optimization performance by solving this benchmark.

The contributions of this research to science are as follows:

- Introduced a new nature-inspired dynamic optimization strategy for ACO algorithm with improved interstate convergence, called ACO with Aphids. The strategy is modelled by mimicking the real-life symbiotic relationship between ants and aphids.
- Provided the description of ACO with Aphids algorithm with enough detail to make this strategy possible to apply to any dynamic optimization problem.
- Tested and proved the superior performance of ACO with Aphids algorithm solving event-triggered Dynamic Multidimensional Knapsack Problem (DMKP) against two most popular competing strategies: Pheromone-Sharing and Full-Restart.

The rest of this paper is structured as follows. Section 2 summarises the Multidimensional Knapsack Problem (MKP) and explains the dynamic variant DMKP. Section 3 explains the observed real-world ants and aphids' symbiotic relationship and then introduces a dynamic optimization strategy model based on this relationship. Detailed ACO with Aphids model includes pseudo-code, flowchart, and general formulas needed to implement the model for a wide range of dynamic optimization problems. Sections 4 and 5 are dedicated to the experimental study of ACO with Aphids solving DMKP. Section 4 describes all problem-specific algorithm parameters, datasets, and measurements used for the experiments. Then Section 5, the results of ACO with Aphids are compared against Full-Restart and Pheromone-Sharing strategies. Finally, conclusions and future directions are presented in Section 6.

## 2 DOP benchmark: Dynamic Multidimensional Knapsack Problem

This research focuses on solving the Dynamic Multidimensional Knapsack Problem (DMKP) due to algorithm applicability to real-world optimization problems and the availability of fully defined DMKP benchmark datasets [46]. DMKP is an extension of the Multidimensional Knapsack Problem (MKP), where dynamic changes occur to item profits, item weights, and knapsack capacities. The dynamic changes of DMKP are not predictable by the optimization algorithm and require a new solution after each dynamic change occurs.

MKP is a well-known academic benchmark optimization problem that has many practical uses like cargo loading [47], layout problem [48], budgeting [49] and portfolio [50] management, to name a few. MKP aims to find a set of items where the combined profit of those items is as high as possible while the combined weight fits in all knapsacks.

$$maximize \sum_{i=1}^{n} x_i \times P_i \quad (1)$$

$$subject\ to \sum_{i=1}^{n}(x_i \times W_{i,k}) \leq C_k, \quad \forall(k)\ where\ k \in (\mathbb{N} \leq m) \quad (2)$$

where $n$ and $m$ are the numbers of items and knapsacks in the problem, respectively. $x_i \in \{0,1\}$ is a binary decision vector to take the item $I_i$. $P_i$ is the profit of the item $I_i$. $W_{i,k}$ is the weight of $i^{th}$ item for the $k^{th}$ knapsack. $C_k$ is the capacity of the $k^{th}$ knapsack. The equations state maximize the profit of the items in the decision vector, subject to item weights not exceeding all corresponding knapsack capacities.

DMKP is also an academic problem that uniquely benefits large-scale real-world optimization problems with some degree of dynamism. The DMKP is a dynamic combinatorial optimization problem, and it is formulated as a sequential series of static MKP instances called states. Between sequential states, the numerical difference of each item's profit, item weights, and knapsack capacities should be reasonably small, indicative of problem dynamism that occurs in the real world. The DMKP problem aims to maximize the total profit of each state before a dynamic change occurs. The result of DMKP is the sum of each state's maximum profit.

$$maximize \sum_{s=0}^{S_{max}} \left( \sum_{i=1}^{n} x_{s,i} \times P_{s,i} \right) \qquad (3)$$

$$subject\ to \sum_{i=1}^{n} (x_{s,i} \times W_{s,i,k}) \leq C_{s,k}, \qquad \forall (s,k)\ where\ k \in (\mathbb{N} \leq m), \qquad s \in (0\ \leq \mathbb{N} \leq S_{max}) \qquad (4)$$

where $s$ is the index of the state, and $S_{max}$ is a number of states in the DMKP problem. The equations state maximize the profit of the items in the decision vector in every DMKP state, subject to item weights not exceeding all corresponding knapsack capacities in every state.

Generally, the degree of dynamism in the literature is defined as the frequency and the magnitude of dynamic change [1]. However, for benchmark dynamic combinatorial optimization problems, the main focus is only on the magnitude of dynamic change. The frequency of dynamic change can be set to a constant time window such that a fair comparison of dynamic optimization strategies could be performed.

## 3 Ant Colony Optimization with Aphids

The original Ant Colony Optimization (ACO) algorithm was described by Dorigo in his doctoral thesis [51], solving Traveling Salesman Problem in 1992. ACO algorithm has been modelled to mimic real ants' behaviour. While navigating, ants deposit pheromone on their path, and then other ants sense it and are drawn to it. A stronger pheromone trail attracts more ants compared to a weaker pheromone trail. When an ant travels a long distance from the food source to the nest, the pheromone trail is naturally spread out over that distance, and the pheromone evaporates faster, making the trail unattractive. On the other hand, if the path is short, each ant's round trip time is shorter, allowing to deposit more pheromone on a short path and attract even more ants. Such ant behaviour is fundamentally iterative and allows ants to explore all the available areas for food sources and exploit already found food sources using the shortest travel path.

However, this optimization algorithm was not modelled with dynamic optimization problems in mind. But luckily, some ants manifest another behaviour, herding aphids. In nature, those ants tend to aphids and collect their produced honeydew, providing an additional source of nutrition along with their usual scavenged food. Within ants' pheromone, aphids behave differently. They move less and produce more honeydew, while ants protect them from predators. Also, aphids rely on ants to relocate them when environmental conditions change. For example, ants move them onto new fresher plants when the plant is no longer fresh. Ants can prey on aphids once aphid's honeydew production decreases or aphids' population is too large.

## 3.1 ACO with Aphids design

We propose to use aphids in ACO algorithm to increase performance of discrete dynamic optimization. In this algorithm, the aphids are modelled as immobile food producers for ants to pick up. Ants must pick up this food at each dynamic change of the problem, creating a baseline pheromone for the new dynamic environment proportional to aphids' distribution around search space. Then ants continue to explore the current state of the environment by optimizing the combinatorial optimization problem. Then after optimization is finished, ants kill a portion of aphids and lay some new aphids on the best edges of the current state of the environment. Aphids do not move independently, but after each dynamic change, ants relocate a portion of aphids to a better location according to precalculated heuristic information.

## 3.2 Optimization system

For discrete dynamic optimization, the ACO algorithm runs within the optimization system to separate concerns of the optimization process and the data. The optimization system handles fully defined benchmark data. The optimization system mimics real-world dynamic optimization scenarios where future dynamic change is unknown. Each state of the dynamic optimization problem is dispatched with a constant time interval for the ACO to solve. Then ACO solves the optimization problem at its current state until the next state is dispatched. The system also records the fitness score and results of the ACO optimization for every state.

## 3.3 ACO with Aphids algorithm

ACO with Aphids design extends Max-Min Ant System (MMAS) introduced by Stützle and Hoos [52]. Aphids represent learned information mediators across all states of the dynamic optimization problem. For each dynamic optimization problem's state, the algorithm initializes search space with all feasible edges, sets each edge's pheromone to the default pheromone level $\tau_0$, and precalculates heuristic information. Before the ant search starts, ants perform two additional steps unique to the ACO with Aphids algorithm. Firstly, they relocate aphids based on new search space heuristic information and aphids' relocation parameter. Then, ants collect honeydew produced by aphids and set the initial pheromone level to base the search in the new environment. Then ants perform iterative search normally for a given time span. Once the search is terminated, ants kill a portion of aphids according to the aphids' kill parameter. And finally, ants lay down new aphids based on the best solution found for the current state and aphids' lay down parameter. The algorithm is further summarised in the pseudo-code below.

*Table 2: ACO with Aphids pseudo-code*

| **Algorithm 1:** ACO with Aphids | |
|---|---|
| **Input:** Dynamic optimization problem dataset | |
| **Output:** Solutions to the dynamic optimization problem | |
| 1. | Initialize Aphids to default aphids' level |
| 2. | **FOR** state **IN** dataset states **DO** |
| 3. |    Load state data to memory |
| 4. |    Prepare search space subroutine |
| 5. |    Initialize Ants' pheromone |
| 4. |    Relocate Aphids |
| 7. |    Collect honeydew by ants |
| 8. |    **WHILE** no events **AND** no termination **DO** |
| 9. |      Build Ant solution subroutine |
| 10. |      Update ants' pheromone |
| 11. |    **END WHILE** |
| 12. |    Record best state's solution |
| 13. |    Kill portion of aphids |
| 14. |    Lay new Aphids based on the best solution |
| 15. | **END FOR** |

### 3.3.1 Initialize Aphids to default aphids' level

Aphids are initialized only once for dynamic optimization, and then aphids evolve together with the dynamic environment. At the very start, aphids are initialized uniformly for the whole search space. The parameter must be above zero for the optimization algorithm to work correctly. The default value is $A_0 = 1$.

$$A_{j,i} := A_0, \quad \forall (j,i) \tag{5}$$

Aphids' level is assigned to the default aphids' level on all edges, where, $A_{j,i}$ is the aphids' level of $j$ node and $i$ edge, $A_0$ is the default aphids' level parameter.

### 3.3.2 Load data to memory and prepare search space

At any given point during dynamic optimization, only one dynamic optimization state is loaded into memory. Previous optimization states are out of date and no longer help in the optimization of the current state. Further states are not revealed to the optimization algorithm as they are technically in the future. Then once the state is revealed, search space has to be prepared for that state. In the prepared search space, each edge has precalculated heuristic information $\eta_{j,i}$. Every optimization problem will have different formulas to calculate heuristic information since this heuristic information is based on the expert knowledge of the optimization problem that provides a myopic guide to the ACO algorithm.

### 3.3.3 Initialize ants' pheromone

In the Max-Min Ant System, static optimization starts with an initial pheromone matrix of default value $\tau_0$ set to each edge. The same principle is applied to dynamic optimization. ACO with Aphids algorithm initializes the pheromone matrix for each optimization state to default pheromone value.

$$\tau_{j,i} := \tau_0, \quad \forall (j,i) \tag{6}$$

Ants' pheromone is assigned to the default ants' pheromone level on all edges, where, $\tau_{j,i}$ is the pheromone level of $j$ node and $i$ edge, $\tau_0$ is the default pheromone level parameter.

### 3.3.4 Relocate aphids

When the optimization system introduces the new state, new heuristic information is precalculated that can roughly point to areas where aphids should be placed. Ants use precalculated heuristic information to relocate aphids from the search space areas with worse heuristic information to areas with a better heuristic. Aphids are partially relocated based on heuristic information because the heuristic information is only a myopic guide to improve initial convergence. Relocation of aphids noted by aphids' relocation parameter $A_r$. The aphids' relocation parameter is a multiplier value and must not be negative $A_r > 0$. The default value of the aphids' relocation parameter $A_r = 1$. However, the parameter value can be higher if heuristic information variance is low, and vice versa.

$$A_{j,i} := A_{j,i} \times (1 + (\eta_{j,i} - \bar{\eta}) \times A_r), \quad \forall (j,i) \tag{7}$$

Aphids' level is reassigned to new aphids' level based on heuristic information and aphids' relocation parameter, where, $A_{j,i}$ is the aphids' level of $j$ node and $i$ edge, $A_r$ is the aphids' relocation parameter, $\eta_{j,i}$ is heuristic information of $j$ node and $i$ edge, and $\bar{\eta}$ is average heuristic information value across the entire search space.

### 3.3.5 Collect honeydew

Honeydew is the crucial product of aphids. Before the iterative search starts, ants pick up the honeydew produced by aphids and lay pheromone on those edges where honeydew is produced. The amount of pheromone laid down is proportional to the amount of honeydew and, in turn, proportional to the number of aphids living on the edge. The default value of the aphids' honeydew production parameter $A_h = 1$. Increasing or decreasing this parameter value increases or decreases aphids' total influence on the dynamic search. However, too much honeydew might affect ants' ability to evaporate excess pheromone created while collecting honeydew.

$$\tau_{j,i} := \tau_{j,i} + (A_{j,i} \times A_h), \quad \forall (j,i) \tag{8}$$

Ants' pheromone is reassigned to a new level based on aphids' level and honeydew production rate, where, $\tau_{j,i}$ is the pheromone level of $j$ node and $i$ edge, $A_{j,i}$ is the number of aphids currently living on the $j$ node and $i$ edge, and $A_h$ is the aphids' honeydew production rate.

This newly laid pheromone forms a solid starting point for ACO to find good initial solutions after the dynamic change has occurred. Aphids' impact on pheromone is applied only once, and its effect does not negatively impact the iterative search convergence. Ants' pheromone normally evaporates as the search progresses, allowing ants to explore search space efficiently without getting stuck in the local optima.

### 3.3.6 Build ant solution

ACO with Aphids performs iterative search normally as described in MMAS [52]. In the iterative search, a set of ants each builds a complete solution independently. Each ant starts the search with an empty partial solution $s_p = \emptyset$. Then the ant searches for a single edge to add to the partial solution. Ant stochastically adds edge to the solution based on the calculated edge's probability without exceeding the optimization problem constraints. The solution is complete once every node's objective is met or constraints are exhausted.

### 3.3.7 Update ants' pheromone

All completed ant solutions are evaluated against fitness function within a single iteration. The best solution is then passed to update the global pheromone. Pheromone update consists of two steps, first evaporation and second lay down. During the evaporation step, a portion of pheromone is reduced by evaporation parameter $\rho$ as in the following equation (9). Then the best ant solution is taken to lay down pheromone on edges that it has visited while building the solution as in the following equation (10):

$$\tau_{j,i} := \tau_{j,i} \times (1 - \rho), \quad \forall (j,i) \tag{9}$$
$$\tau_{j,i} := \tau_{j,i} + \rho \times \Delta\tau_0, \quad \forall (j,i) \in s_p \tag{10}$$

where $\rho$ is a constant parameter of the pheromone evaporation rate introduced by Dorigo and Stützle [53], $\Delta\tau_0$ is the pheromone update rate, $s_p$ is the solution of the chosen ant to lay down the pheromone.

### 3.3.8 Termination criteria

The search runtime is terminated of a current dynamic state when either one of three conditions occurs, event-triggered, time-based or iteration based. The first condition is when the dynamic change event is triggered. The system mimics the real-world dynamic change in the optimization problem's instance and dispatches new and updated problem search space. The system treats the solutions to the old dynamic state as invalid to the new search space and restarts the search. The second condition is time-based. The optimization is considered complete when the optimization time reaches a predefined time limit allocated for the optimization. The third condition is iteration based. Similarly to time-based termination, the optimization is considered complete when the iterative search has performed a predefined number of iterations.

### 3.3.9 Kill portion of aphids

After the iterative search is finished, a portion of aphids is killed based on the aphids' kill rate parameter $A_k$. Killing aphids procedure ensures that aphids' population does not grow too much. Also, the killing aphids procedure further removes aphids from poorly performing edges throughout multiple optimization states. Like in nature, a portion of aphids must die such that aphids populations stays in equilibrium $A_r > 0$. The default value of the aphids' kill rate parameter $A_k = 0.5$, which kills half of the remaining aphids.

$$A_{j,i} := A_{j,i} \times (1 - A_k), \qquad \forall (j, i) \tag{11}$$

Aphids' level is reassigned to a new aphids' level based on aphids' kill rate, where, $A_{j,i}$ is the number of aphids on $j$ node and $i$ edge, and $A_k$ is the aphids' kill rate.

### 3.3.10 Lay new aphids

The best solution found by ACO for the state is representative of that state's environment. Ants lay down new aphids on the edges of the search space that are included in the best solution. Lay down aphids procedure increases the number of aphids on well-performing edges and strengthens honeydew production. The default value of the aphids' lay down parameter $A_l = 1$. However, this parameter can be higher if typical solutions contain a small portion of edges available in the search space and vice versa.

$$A_{j,i} := A_{j,i} + A_l, \qquad \forall (j, i) \in s_p \tag{12}$$

Aphids' level is reassigned to a new aphids' level based on aphids' lay down rate, where $A_{j,i}$ is the number of aphids, $A_l$ is the aphids' lay down rate, and $s_p$ is the best solution chosen for lying down the aphids.

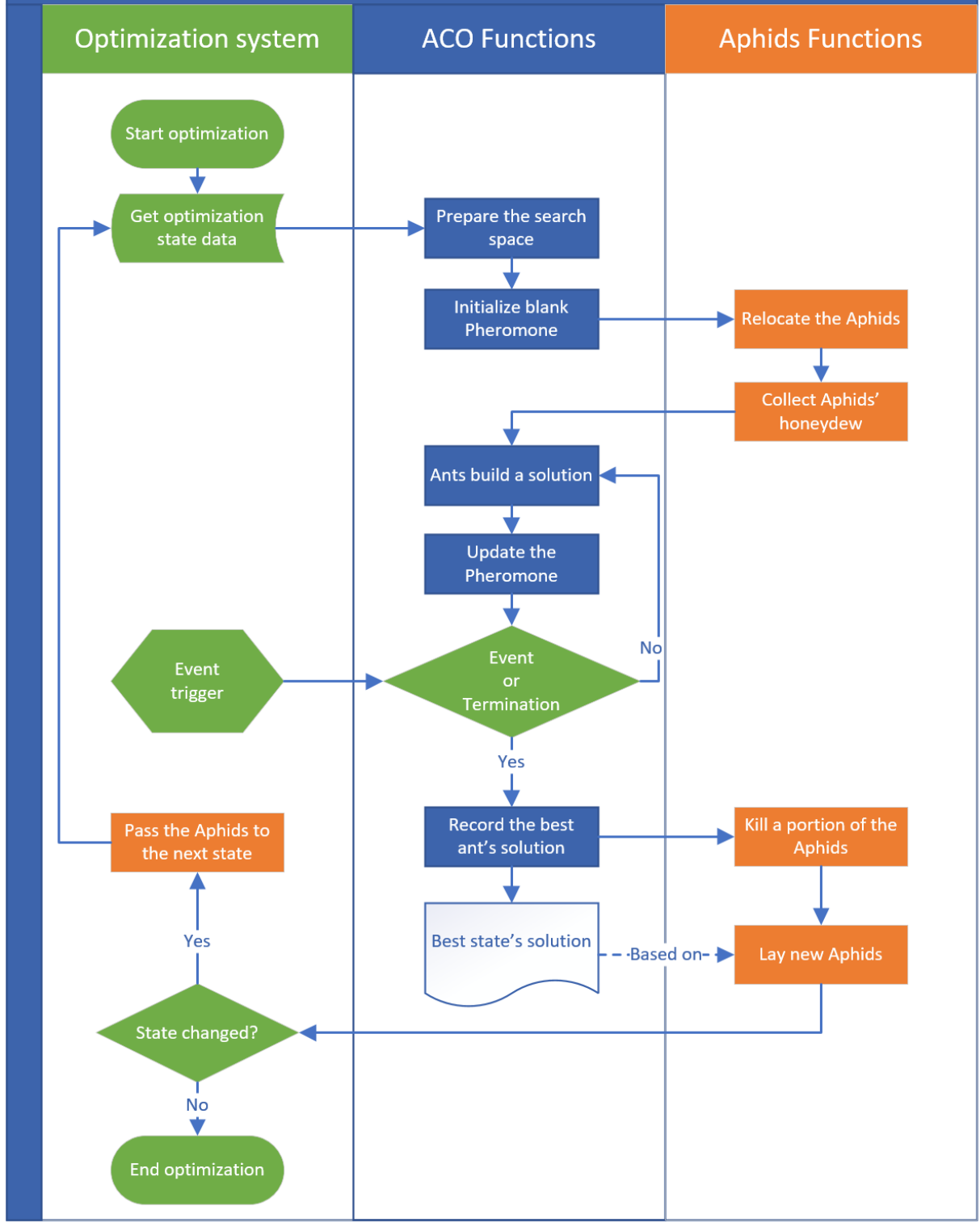

*Figure 9: ACO with Aphids algorithm flowchart. The green colour represents Optimization system steps, the blue colour represents ACO algorithm steps, and the orange colour represents novel steps to ACO with Aphids algorithm.*

Overall, ACO with Aphids dynamic optimization strategy offers a new method to share information between dynamic states. The inter-state information sharing is mediated by a new passive agent called Aphids. Aphids' primary purpose is to produce honeydew which is then turned into the pheromone before the main ACO search starts. The aphids are distinct from the pheromone because aphids have different update logic from the pheromone. The main advantage of using ACO with Aphids is that the newly initialized pheromone before each state's optimization efficiently combines learned information through the whole optimization with the heuristic of the current optimization state. On the other hand, the algorithm might not perform better than the standard Pheromone-Sharing strategy when the optimization problem does not have reliable heuristic information. Furthermore, Aphids add a negligible computational complexity increase, similar to a single iteration pheromone update with linear complexity $O(n)$, because aphids are only used before and after the ACO search, which is minimal compared to the ACO search with quadratic complexity $O(n^2)$.

## 4 Experimental setup

The experimental work presents the comparison of the proposed ACO with Aphids dynamic optimization strategy against the two most popular ACO dynamic optimization strategies: Full-Restart and Pheromone-Sharing [34] [54] [55].

### 4.1 Experimental dataset

A fully defined Dynamic Multidimensional Knapsack Problem (DMKP) benchmark is chosen to test the proposed algorithm performance. Each benchmark dataset includes complete information about each optimization problem instance called states [46] [56] [57]. Each benchmark dataset contains 101 states, where the first state is the initial state based on the static benchmark instance of the MKP and 100 deterministically generated states. This benchmark is perfect for reliably testing an event-triggered dynamic optimization system and comparing the results of dynamic optimization algorithms. Also, DMKP suits this aim well because algorithmic solutions solving MKP have a wide range of possible real-world applications.

From this benchmark library, 55 datasets of the GK group are used for the experimental work. Each GK dataset has a unique complexity and dynamism combination. There are 11 complexity levels in the range from the GK01 dynamic dataset group with 100 items and 15 knapsacks to the GK11 dynamic dataset group with 2,500 items and 100 knapsacks, and 5 dynamism levels indicated by the State Adjustment Magnitude $\Delta$ parameter: $\Delta = 0.01, \Delta = 0.02, \Delta = 0.05, \Delta = 0.1, \Delta = 0.2$. The $\Delta$ parameter is also called SAM in the code and charts with no special characters' support. For the first experiment of ACO with Aphids hyper-parameter tuning, only GK03 and GK08 complexities are used with all 5 dynamism levels. All 11 dataset groups from the GK group with all five dynamism levels are used for the second experiment of a full comparison of dynamic optimisation strategies.

### 4.2 Baseline ACO algorithm and optimization system

All three dynamic optimization strategies are implemented within the same baseline ACO core algorithm, solving static MKP benchmarks [58]. High-quality MKP solutions were possible to achieve by utilizing a dynamic impact evaluation method in the ACO edge's probability calculation. Also, this baseline ACO algorithm implementation efficiently utilizes modern multicore computer architectures by running multiple ant searches in parallel within one iteration and synchronising before the pheromone update.

ACO algorithm parameters have been tuned in previous research [58] and used throughout all experimentation. The best combination of pheromone parameters are: $\tau_{max} = 1$, $\tau_{min} = 0.001$, $\tau_0 = 1$, $\Delta\tau_0 = 1$, $\rho = 0.1$. Configuration of probability parameters: $\alpha = 1$. Configuration of probability parameters: $\alpha = 1, \beta = 0, \gamma = 8, q0 = 0.01$. Each iteration runs 512 ants in parallel.

The heuristic information of each item $\eta_i$ used in this algorithm for DMKP represents the relative value of the item for a given state as a function of the item's profit over the total weight of the item:

$$\eta_i = \frac{P_i}{W_i} \tag{13}$$

where, $P_i$ is the profit if the item and $W_i$ is the total weight of the item. It is important to note that the heuristic information is calculated for a given discrete dynamic optimization state. Each state may have items with different profits and weights.

A fair comparison of dynamic optimization algorithm performance is ensured by the optimization system dispatching an event-trigger at a precise interval based on the complexity of the optimization problem. Event-trigger dispatch based on time eliminates any variations in algorithm execution overheads related to strategy. The optimization system has the rule to dispatch events after a time period proportional to the number of items in the dataset. Each state has 1 second to execute optimization for every 200 items in the dataset. For example, any dataset in the GK01 dataset group has 100 items. Therefore, each state is allowed to execute for only 0.5 seconds. The GK11 is the largest dataset group with 2500 items per dataset. Therefore, each state is allowed to execute for 12.5 seconds.

The experimental environment is also carefully controlled to ensure the consistency of computational power among the tests. All tests have been executed on the AMD Threadripper 2990WX system with the clock running at 2.9Ghz. Only one experiment has been allowed to run on the system simultaneously without any other background tasks. Execution parallelism is set to 32 threads on the first NUMA node. The ACO core algorithm and dynamic optimization strategies are implemented in C++ language and compiled with the MSVC compiler.

### 4.3 Experimental measurements

The absolute performance of the DMKP benchmark instance is the sum of each state's result profits. The total profit sum is a good overall performance indication because each state equally contributes to the final result. A profit improvement of any state's result directly improves the total profit. The result performance is only comparable against the same benchmark instance because every dataset instance will have a different optimal total profit value. However, if the optimal or best-know values are available, a comparable metric can be calculated as the result's percentage difference from the best-known result called the "result gap". The goal is to minimize the result gap, ideally to zero, which means the best-known or the optimal solution is found. The result gap can be calculated for any single solution and plotted on a graph. Also, dynamic algorithm performance can be measured using the difference in the result gap that occurs after the dynamic change called the "gap slip". Although gap slip is not a primary objective of the optimization, it indicates how well the optimization algorithm tackles a dynamic problem and should be minimized.

This research performs quantitative analysis across all benchmark instances using "result gap" and "gap slip" metrics. The average result gap and average gap slip values of each dynamic benchmark dataset are calculated from all dynamic states, as shown in Figure 10. The best-known results are taken from a verified public repository [59]. Furthermore, all presented experimental measurements are the averages taken from 10 algorithm runs, and the standard deviation of the result gap is calculated to prove statistical significance.

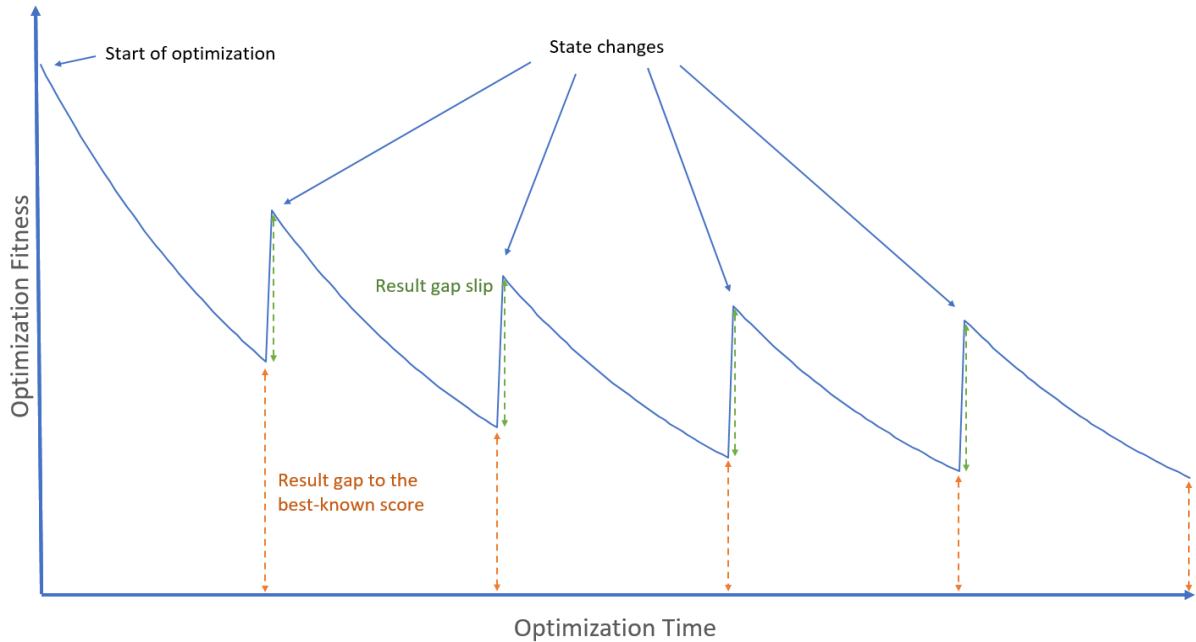

*Figure 10: Experimental measurements visualization. The orange line represents a measurement of each state's result gap to the best know profit score. The green line represents the result gap slip after the dynamic change. The total dynamic optimization result gap is an average of all states' result gap, and the total dynamic optimization result gap is an average of all states' gap slip.*

## 5   Experimental results

The experimental work is split into two parts. The first part is dedicated to an iterative tuning of ACO with Aphids strategy hyper-parameters using a reduced benchmark dataset sample. Then the second part is dedicated to comparing ACO with Aphids strategy with Full-Restart and Pheromone-Sharing strategies.

### 5.1   ACO with Aphids hyper-parameter tuning results

ACO with Aphids algorithm presents with four new Aphids functions with tuneable parameters, as shown in Figure 9. These parameters must be carefully tuned to maximize the performance of proposed algorithm. Meanwhile, Full-Restart and Pheromone-Sharing strategies do not have additional functions with tuneable parameters, and all shared hyper-parameters have been already tuned as part of prior research [58]. Initially, all Aphids tuneable parameters are initialized to the default values, as shown in Table 3, and sweep the tested the algorithm's performance by varying one parameter value per test. The tests are performed incrementally. The first test starts with all default parameter values, and the following tests use the best parameter values found in the previous tests. Hyper-parameter tuning is general to the algorithm, and tests can be performed on a subset of datasets to reduce computational demand. GK03 and GK08 dataset groups with all 5 dynamism levels are selected for the hyper parameter tuning test to represent low and high complexity members of the benchmark.

Table 3: Aphids' tuneable parameters table. Each parameter has a default value, min-max value range used in tests, and test resolution.

| Parameter | Default Value | Min Test Value | Max Test Value | Test Resolution |
|---|---|---|---|---|
| Aphids' relocation: $A_r$ | 1 | 0 | 2 | 0.25 |
| Aphids' honeydew production: $A_h$ | 1 | 0 | 2 | 0.25 |
| Aphids' lay down rate: $A_l$ | 1 | 0 | 2 | 0.25 |
| Aphids' kill rate: $A_k$ | 0.5 | 0.01 | 1 | 0.2 |

The first tested parameter is Aphids' relocation $A_r$. This parameter partially allows Ants to relocate a portion of the Aphids based on the heuristic information. The heuristic information is not a perfect measure of fitness. However, it can prove helpful for dynamic optimization. The results show a clear optimization improvement with higher $A_r$ values and best-tested configuration is with $A_r = 2$.

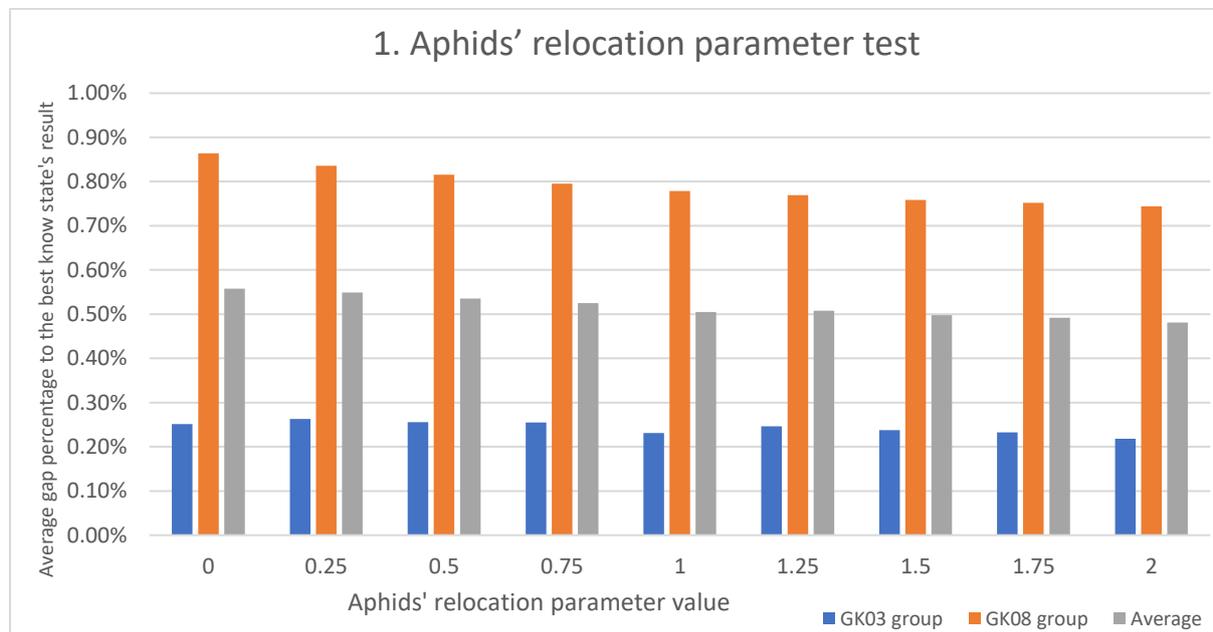

Figure 11: ACO with Aphids hyper-parameter tuning test number 1. Aphids' relocation parameter test. The results show the best dynamic optimization performance is achieved using $A_r = 2$.

The second tested parameter is Aphids' honeydew production $A_h$. This parameter controls how much honeydew each aphid produces at the start of the iterative search, where higher values result in stronger pheromone trails at the start of each dynamic optimization state. The results show a significant optimization improvement with higher $A_h$ values up to $A_h = 1$ after which the results had no more improvement. With honeydew production above 1, Aphids saturate Ants' pheromone trails with honeydew, and extra honeydew does not further benefit the dynamic search.

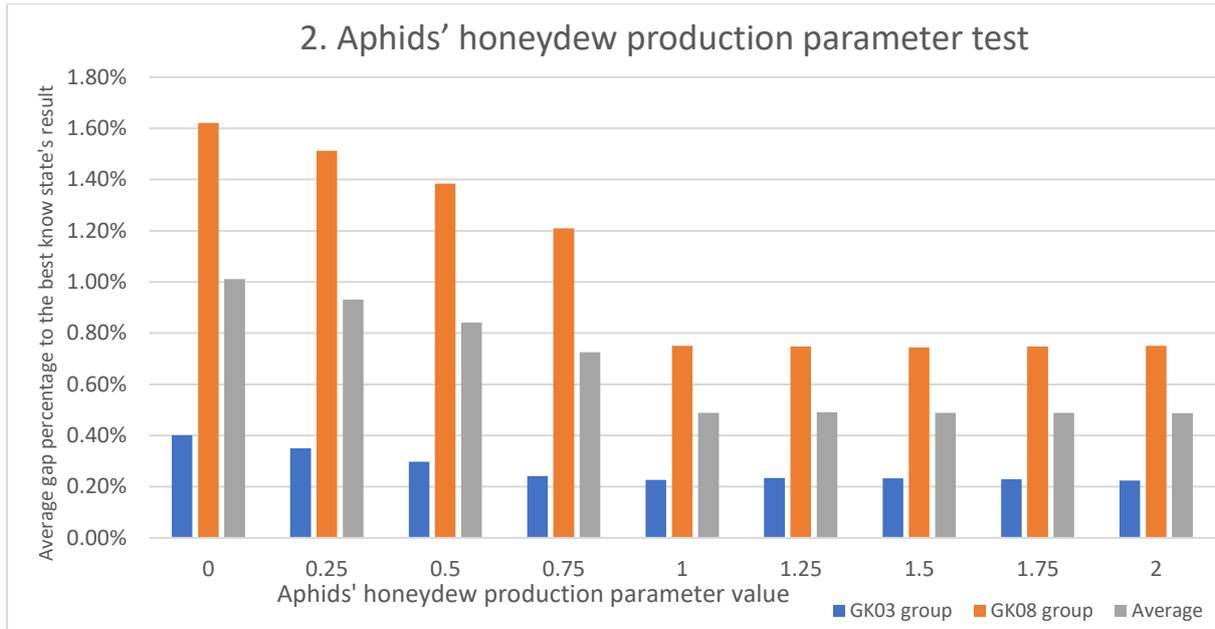

Figure 12: ACO with Aphids hyper-parameter tuning test number 2. Aphids' honeydew production parameter test. The results show the best dynamic optimization performance is achieved using $A_h = 1$.

The third tested parameter is Aphids' lay down rate $A_l$. This parameter controls how much Ants lay down new Aphids on the edges of the best solution obtained in the previous state's optimization. The results show significant improvement with a lay down rate above 0, which is to be expected. When no aphids are laid, the information is no longer shared between states. When parameter $A_l = 1$, it appears to be a middle ground where the smaller complexity GK03 group is not significantly diminished, and the larger complexity GK08 group performs well too.

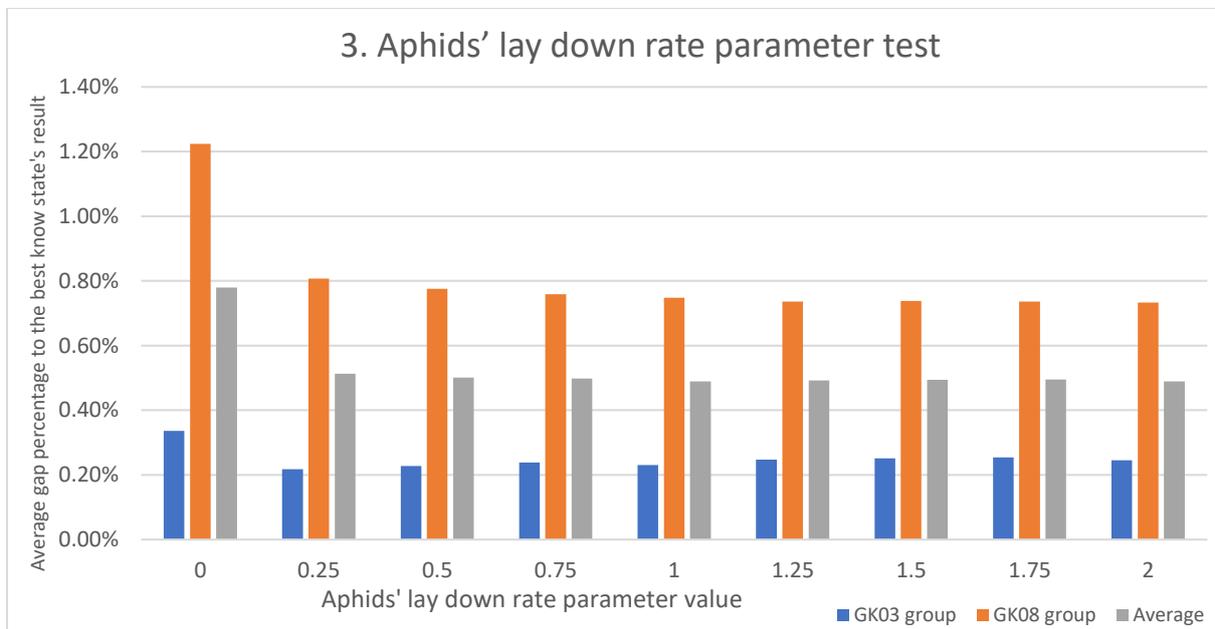

Figure 13: ACO with Aphids hyper-parameter tuning test number 3. Aphids' lay down rate parameter test. The results show the best dynamic optimization performance is achieved using $A_l = 1$.

Finally, the last parameter tested is the Aphids' kill rate $A_k$. This parameter controls which portion of the aphids is killed after completing the iterative search. This procedure ensures that the population does not grow too much and aphids are removed from poorly performing edges. The results show a preference for a higher kill rate $A_k = 0.8$, but not killing all aphids, which leaves a small portion of carry-over aphids from previous states.

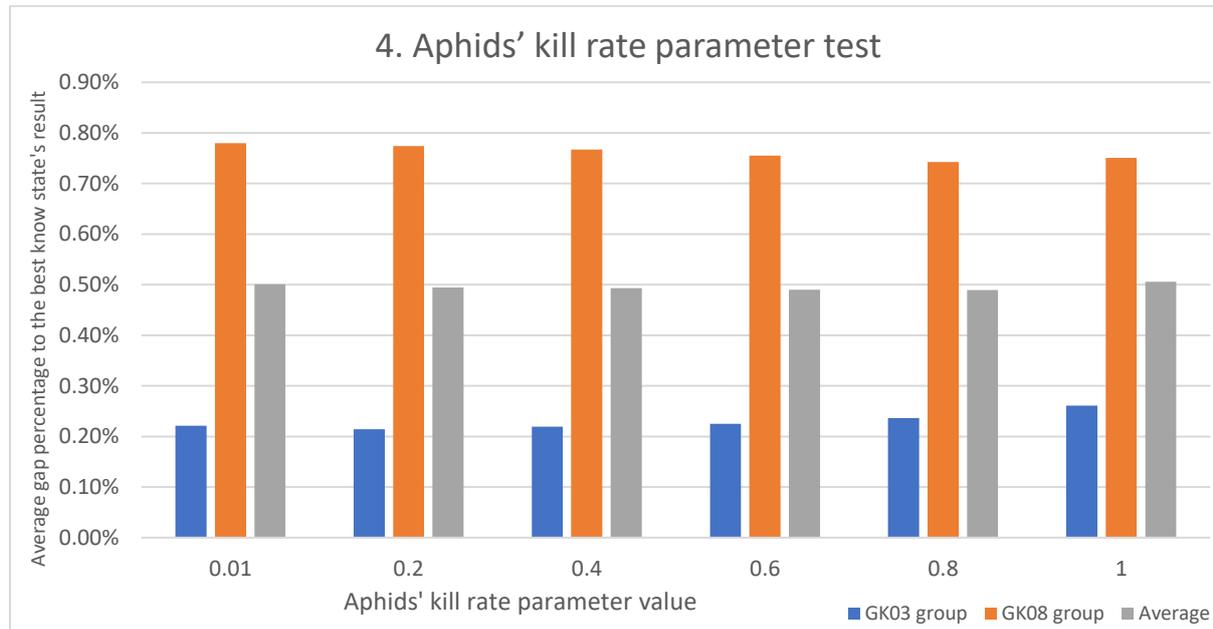

*Figure 14: ACO with Aphids hyper-parameter tuning test number 4. Aphids' kill rate parameter test. The results show the best dynamic optimization performance is achieved using $A_k = 0.8$.*

In summary of ACO with Aphids hyper-parameter tuning, the Aphids specific hyper-parameters have been tested to find the best configuration to optimize the DMKP problem. The configuration: $A_r = 2$, $A_h = 1$, $A_l = 1$, $A_k = 0.8$ has been shown to perform the best. This ACO with Aphids hyper-parameter configuration will be used for the next experiment comparing ACO with aphids against the other two most popular ACO dynamic optimization strategies.

## 5.2 ACO with Aphids comparison with other ACO dynamic optimization strategies result

This experiment is dedicated to a full comparison of ACO with Aphids dynamic optimization algorithm strategy with the other two most common dynamic optimization strategies for ACO algorithm, Full-Restart and Pheromone-Sharing strategies. All strategies are tested using 55 fully-defined DMKP benchmark datasets with 11 dataset complexity groups from smallest GK01 to largest GK11, and 5 dynamism levels from SAM-0.01 to SAM-0.2. All three dynamic optimization strategies are implemented on the same ACO search core, and the experiments are conducted on the isolated test system. Therefore, the only test variables are the dynamic optimization strategies. The statistical results' significance is proven with acquired each data point by running the dynamic optimization ten times.

In Table 4, the results show an average result gap of each dynamic optimization strategy for every dynamic benchmark dataset. The table also includes the average summary per dataset group and dynamism level, as well as the total average of the dynamic optimization strategy. Overall, the ACO with Aphids has shown the best performance, with an average result gap of 0.519%. The second best was the Pheromone-Sharing strategy, with an average gap of 0.733%. Lastly, the Full-Restart strategy achieved an average result gap of 1.092%. In relative terms, these results show that the ACO with Aphids strategy performed 110% better than the Full-Restart strategy and 41% better than the Pheromone-Sharing strategy.

Table 4: Dynamic optimization average result gap of all optimization strategies. Each data point is an average of all dynamic states' result gap over ten algorithm runs. (Lower is better)

| Average results gap | | Dynamism level | Dataset group | | | | | | | | | | | Average |
|---|---|---|---|---|---|---|---|---|---|---|---|---|---|---|
| | | | GK01 | GK02 | GK03 | GK04 | GK05 | GK06 | GK07 | GK08 | GK09 | GK10 | GK11 | |
| Strategy | ACO with Aphids | SAM-0.01 | 0.158% | 0.160% | 0.117% | 0.171% | 0.124% | 0.159% | 0.339% | 0.429% | 0.694% | 0.652% | 0.563% | 0.324% |
| | | SAM-0.02 | 0.160% | 0.191% | 0.125% | 0.160% | 0.139% | 0.189% | 0.294% | 0.437% | 0.640% | 0.696% | 0.694% | 0.339% |
| | | SAM-0.05 | 0.215% | 0.259% | 0.173% | 0.241% | 0.177% | 0.335% | 0.355% | 0.595% | 0.756% | 0.902% | 0.927% | 0.449% |
| | | SAM-0.1 | 0.283% | 0.347% | 0.233% | 0.454% | 0.322% | 0.539% | 0.553% | 0.951% | 0.969% | 1.208% | 1.142% | 0.636% |
| | | SAM-0.2 | 0.282% | 0.407% | 0.281% | 0.566% | 0.391% | 0.708% | 0.900% | 1.354% | 1.413% | 1.641% | 1.371% | 0.847% |
| | | Average | 0.220% | 0.273% | 0.186% | 0.319% | 0.231% | 0.386% | 0.488% | 0.753% | 0.894% | 1.020% | 0.939% | **0.519%** |
| | Full-Restart | SAM-0.01 | 0.142% | 0.199% | 0.481% | 0.933% | 0.867% | 1.100% | 1.622% | 1.777% | 2.231% | 2.011% | 1.661% | 1.184% |
| | | SAM-0.02 | 0.124% | 0.191% | 0.425% | 0.876% | 0.766% | 1.040% | 1.520% | 1.822% | 2.285% | 2.266% | 1.890% | 1.200% |
| | | SAM-0.05 | 0.149% | 0.194% | 0.373% | 0.799% | 0.630% | 0.969% | 1.399% | 1.649% | 2.234% | 2.134% | 1.584% | 1.101% |
| | | SAM-0.1 | 0.166% | 0.177% | 0.329% | 0.735% | 0.655% | 0.916% | 1.254% | 1.504% | 1.985% | 1.896% | 1.358% | 0.998% |
| | | SAM-0.2 | 0.127% | 0.163% | 0.325% | 0.683% | 0.590% | 0.888% | 1.198% | 1.547% | 1.876% | 1.915% | 1.433% | 0.977% |
| | | Average | 0.142% | 0.185% | 0.387% | 0.805% | 0.701% | 0.983% | 1.399% | 1.660% | 2.122% | 2.044% | 1.585% | **1.092%** |
| | Pheromone-Sharing | SAM-0.01 | 0.180% | 0.196% | 0.211% | 0.236% | 0.184% | 0.263% | 0.227% | 0.454% | 0.750% | 0.882% | 1.137% | 0.429% |
| | | SAM-0.02 | 0.190% | 0.219% | 0.218% | 0.260% | 0.190% | 0.364% | 0.268% | 0.529% | 0.805% | 1.050% | 1.229% | 0.484% |
| | | SAM-0.05 | 0.276% | 0.332% | 0.283% | 0.345% | 0.225% | 0.486% | 0.420% | 0.851% | 1.082% | 1.271% | 1.300% | 0.625% |
| | | SAM-0.1 | 0.337% | 0.453% | 0.377% | 0.612% | 0.381% | 0.799% | 0.831% | 1.419% | 1.938% | 1.720% | 1.344% | 0.928% |
| | | SAM-0.2 | 0.340% | 0.511% | 0.471% | 0.819% | 0.497% | 1.062% | 1.565% | 1.933% | 2.490% | 2.056% | 1.457% | 1.200% |
| | | Average | 0.265% | 0.342% | 0.312% | 0.454% | 0.295% | 0.595% | 0.662% | 1.037% | 1.413% | 1.396% | 1.293% | **0.733%** |

Furthermore, the result gap standard deviation of each experiment is shown in Table 5 to disprove a null hypothesis. Out of all tested dynamic optimization strategies, the Full-Restart strategy had the lowest overall standard deviation of just 0.0104%, the ACO with Aphids strategy had a larger overall standard deviation of 0.0195%, and the Pheromone-Sharing strategy had the largest overall standard deviation of 0.0330%. Statistical significance and rejection of the null hypothesis can be proved using the two-sample unpaired t-test considering the result magnitude, standard deviation, and sample size [60]. ACO with Aphids and Full-Restart strategies sample separations result in a T-value of 81.9 and a P-value $< 1^{-6}$. ACO with Aphids and Pheromone-Sharing strategies sample separations result in a T-value of 17.7 and a P-value $< 1^{-6}$. Both t-test groups reject the null hypothesis and show the exceptionally statistically significant sample separations.

Table 5: Dynamic optimization result gap standard deviation of all optimization strategies. Each data point is a standard deviation of the dynamic optimization result gap with a sample size of 10 runs.

| Result gap standard deviation | | Dynamism level | Dataset group | | | | | | | | | | | Average |
|---|---|---|---|---|---|---|---|---|---|---|---|---|---|---|
| | | | GK01 | GK02 | GK03 | GK04 | GK05 | GK06 | GK07 | GK08 | GK09 | GK10 | GK11 | |
| Strategy | ACO with Aphids | SAM-0.01 | 0.032% | 0.030% | 0.022% | 0.034% | 0.020% | 0.025% | 0.012% | 0.015% | 0.046% | 0.012% | 0.010% | 0.026% |
| | | SAM-0.02 | 0.019% | 0.027% | 0.017% | 0.018% | 0.014% | 0.021% | 0.022% | 0.013% | 0.012% | 0.014% | 0.008% | 0.018% |
| | | SAM-0.05 | 0.020% | 0.027% | 0.012% | 0.022% | 0.011% | 0.019% | 0.012% | 0.012% | 0.022% | 0.019% | 0.012% | 0.018% |
| | | SAM-0.1 | 0.010% | 0.020% | 0.020% | 0.012% | 0.014% | 0.029% | 0.009% | 0.012% | 0.012% | 0.011% | 0.009% | 0.016% |
| | | SAM-0.2 | 0.017% | 0.025% | 0.015% | 0.010% | 0.011% | 0.046% | 0.014% | 0.013% | 0.007% | 0.011% | 0.007% | 0.019% |
| | | Average | 0.021% | 0.026% | 0.017% | 0.021% | 0.015% | 0.030% | 0.014% | 0.013% | 0.024% | 0.014% | 0.009% | **0.020%** |
| | Full-Restart | SAM-0.01 | 0.008% | 0.005% | 0.011% | 0.011% | 0.017% | 0.014% | 0.007% | 0.006% | 0.007% | 0.004% | 0.003% | 0.009% |
| | | SAM-0.02 | 0.008% | 0.009% | 0.011% | 0.012% | 0.019% | 0.008% | 0.006% | 0.005% | 0.010% | 0.006% | 0.003% | 0.010% |
| | | SAM-0.05 | 0.007% | 0.010% | 0.010% | 0.015% | 0.026% | 0.006% | 0.006% | 0.008% | 0.009% | 0.007% | 0.002% | 0.011% |
| | | SAM-0.1 | 0.007% | 0.008% | 0.011% | 0.016% | 0.024% | 0.014% | 0.011% | 0.010% | 0.004% | 0.005% | 0.002% | 0.012% |
| | | SAM-0.2 | 0.010% | 0.012% | 0.008% | 0.010% | 0.019% | 0.006% | 0.008% | 0.006% | 0.007% | 0.006% | 0.004% | 0.010% |
| | | Average | 0.008% | 0.009% | 0.010% | 0.013% | 0.021% | 0.010% | 0.008% | 0.007% | 0.008% | 0.006% | 0.003% | **0.010%** |
| | Pheromone-Sharing | SAM-0.01 | 0.036% | 0.041% | 0.035% | 0.047% | 0.019% | 0.055% | 0.014% | 0.019% | 0.102% | 0.026% | 0.022% | 0.045% |
| | | SAM-0.02 | 0.020% | 0.026% | 0.040% | 0.040% | 0.020% | 0.041% | 0.019% | 0.028% | 0.021% | 0.042% | 0.016% | 0.030% |
| | | SAM-0.05 | 0.023% | 0.026% | 0.016% | 0.028% | 0.012% | 0.055% | 0.021% | 0.044% | 0.018% | 0.057% | 0.012% | 0.032% |
| | | SAM-0.1 | 0.017% | 0.030% | 0.018% | 0.035% | 0.007% | 0.019% | 0.025% | 0.029% | 0.029% | 0.054% | 0.005% | 0.028% |
| | | SAM-0.2 | 0.014% | 0.023% | 0.025% | 0.023% | 0.013% | 0.034% | 0.029% | 0.032% | 0.023% | 0.052% | 0.006% | 0.027% |
| | | Average | 0.023% | 0.030% | 0.028% | 0.036% | 0.015% | 0.043% | 0.022% | 0.031% | 0.050% | 0.048% | 0.014% | **0.033%** |

The average gap slip results show how much performance decreases on average after each dynamic change. The gap slip is calculated using the first iteration's result of the current state minus the last iteration result before the dynamic change has occurred. The average gap slip measurements are shown in Table 6 of each dynamic optimization strategy for every dynamic benchmark dataset. Interestingly, the Pheromone-Sharing strategy achieved the best average gap slip of only 0.240%. Meanwhile, the ACO with Aphids strategy has achieved an average gap slip of 0.304%. Lastly, the Full-Restart strategy achieved the worst average gap slip of 1.420%.

Table 6: Dynamic optimization average gap slip of all optimization strategies. Each data point is an average of all dynamic states gap slip over ten algorithm runs. (Lower is better)

| | Average gap slip | Dynamism level | GK01 | GK02 | GK03 | GK04 | GK05 | GK06 | GK07 | GK08 | GK09 | GK10 | GK11 | Average |
|---|---|---|---|---|---|---|---|---|---|---|---|---|---|---|
| Strategy | ACO with Aphids | SAM-0.01 | 0.050% | 0.073% | 0.103% | 0.125% | 0.155% | 0.167% | 0.128% | 0.108% | 0.013% | 0.014% | 0.005% | 0.086% |
| | | SAM-0.02 | 0.101% | 0.134% | 0.169% | 0.177% | 0.185% | 0.192% | 0.146% | 0.121% | 0.038% | 0.027% | 0.009% | 0.118% |
| | | SAM-0.05 | 0.263% | 0.288% | 0.339% | 0.339% | 0.292% | 0.304% | 0.197% | 0.182% | 0.073% | 0.043% | 0.026% | 0.213% |
| | | SAM-0.1 | 0.563% | 0.587% | 0.607% | 0.553% | 0.532% | 0.506% | 0.334% | 0.307% | 0.178% | 0.121% | 0.043% | 0.394% |
| | | SAM-0.2 | 1.062% | 1.090% | 1.116% | 0.877% | 0.956% | 0.759% | 0.684% | 0.475% | 0.410% | 0.273% | 0.074% | 0.707% |
| | | Average | 0.408% | 0.434% | 0.467% | 0.414% | 0.424% | 0.386% | 0.298% | 0.239% | 0.142% | 0.096% | 0.031% | **0.304%** |
| | Full-Restart | SAM-0.01 | 2.730% | 2.607% | 2.400% | 1.785% | 2.049% | 1.465% | 1.333% | 0.685% | 0.474% | 0.232% | 0.060% | 1.438% |
| | | SAM-0.02 | 2.645% | 2.566% | 2.390% | 1.805% | 2.220% | 1.486% | 1.474% | 0.827% | 0.542% | 0.260% | 0.078% | 1.481% |
| | | SAM-0.05 | 2.518% | 2.528% | 2.337% | 1.704% | 2.101% | 1.407% | 1.700% | 0.904% | 0.709% | 0.349% | 0.071% | 1.484% |
| | | SAM-0.1 | 2.355% | 2.369% | 2.174% | 1.550% | 1.916% | 1.290% | 1.608% | 0.816% | 0.698% | 0.318% | 0.061% | 1.378% |
| | | SAM-0.2 | 2.242% | 2.248% | 2.085% | 1.552% | 1.861% | 1.254% | 1.467% | 0.786% | 0.641% | 0.308% | 0.061% | 1.319% |
| | | Average | 2.498% | 2.464% | 2.277% | 1.679% | 2.029% | 1.380% | 1.516% | 0.804% | 0.613% | 0.293% | 0.066% | **1.420%** |
| | Pheromone-Sharing | SAM-0.01 | 0.041% | 0.052% | 0.044% | 0.047% | 0.054% | 0.043% | 0.063% | 0.077% | 0.042% | 0.043% | 0.027% | 0.048% |
| | | SAM-0.02 | 0.090% | 0.096% | 0.080% | 0.084% | 0.083% | 0.079% | 0.067% | 0.082% | 0.048% | 0.047% | 0.032% | 0.072% |
| | | SAM-0.05 | 0.204% | 0.203% | 0.199% | 0.208% | 0.211% | 0.181% | 0.117% | 0.133% | 0.077% | 0.071% | 0.039% | 0.149% |
| | | SAM-0.1 | 0.477% | 0.476% | 0.466% | 0.434% | 0.461% | 0.365% | 0.263% | 0.226% | 0.146% | 0.119% | 0.043% | 0.316% |
| | | SAM-0.2 | 1.179% | 1.044% | 0.995% | 0.707% | 0.895% | 0.555% | 0.565% | 0.314% | 0.247% | 0.178% | 0.054% | 0.612% |
| | | Average | 0.398% | 0.374% | 0.357% | 0.296% | 0.341% | 0.244% | 0.215% | 0.166% | 0.112% | 0.092% | 0.039% | **0.240%** |

Additionally, Table 7 shows the average result scores of each dynamic optimization strategy for every dynamic benchmark dataset. Each result shows the average total profit summed up for all dynamic states.

Table 7: Dynamic optimization average result scores of all optimization strategies. Each data point is a profit sum of all dynamic states, results averaged over ten algorithm runs and rounded to the nearest integer. (Higher is better)

| | Average results | Dynamism level | GK01 | GK02 | GK03 | GK04 | GK05 | GK06 | GK07 | GK08 | GK09 | GK10 | GK11 |
|---|---|---|---|---|---|---|---|---|---|---|---|---|---|
| Strategy | ACO with Aphids | SAM-0.01 | 46801741 | 49135239 | 70354370 | 71466403 | 93949903 | 95230295 | 238659912 | 232803226 | 716888762 | 707504272 | 1175759342 |
| | | SAM-0.02 | 47010402 | 49076204 | 70365325 | 71420475 | 94271720 | 95439550 | 240011435 | 233887857 | 721364124 | 710927951 | 1180292156 |
| | | SAM-0.05 | 47470009 | 48695642 | 70585120 | 70906032 | 94340404 | 94530635 | 241712888 | 233989576 | 727056981 | 713158456 | 1173446913 |
| | | SAM-0.1 | 47729583 | 49150310 | 70777571 | 70299599 | 94867060 | 93739900 | 242435794 | 233219992 | 726216931 | 712305189 | 1174160002 |
| | | SAM-0.2 | 47967074 | 48184505 | 72238940 | 71412035 | 96706701 | 95166924 | 241527906 | 238498122 | 725575434 | 719242678 | 1192642716 |
| | Full-Restart | SAM-0.01 | 46809064 | 49115980 | 70097599 | 70921311 | 93251309 | 94332748 | 235587148 | 229649572 | 705795787 | 697819782 | 1162772717 |
| | | SAM-0.02 | 47027432 | 49076358 | 70154547 | 70908236 | 93679842 | 94626066 | 237061152 | 230633971 | 709418069 | 699678042 | 1166073879 |
| | | SAM-0.05 | 47501035 | 48727273 | 70443659 | 70509374 | 93912582 | 93929514 | 239180097 | 231507127 | 716230233 | 704285059 | 1165662684 |
| | | SAM-0.1 | 47785816 | 49234358 | 70709165 | 70100749 | 94549864 | 93384977 | 240727335 | 231918996 | 718767177 | 707345303 | 1171591161 |
| | | SAM-0.2 | 48041810 | 48302173 | 72207181 | 71328331 | 96514015 | 94994788 | 240800089 | 238033219 | 722172400 | 717243986 | 1191895429 |
| | Pheromone-Sharing | SAM-0.01 | 46791573 | 49117323 | 70287644 | 71420026 | 93893832 | 95131510 | 238928857 | 232744530 | 716487598 | 705871793 | 1168974104 |
| | | SAM-0.02 | 46996016 | 49062640 | 70300023 | 71348791 | 94223378 | 95272190 | 240076472 | 233671403 | 720169813 | 708397349 | 1173939612 |
| | | SAM-0.05 | 47440660 | 48660046 | 70507538 | 70832201 | 94294890 | 94387154 | 241555504 | 233386074 | 724670231 | 710501659 | 1169028451 |
| | | SAM-0.1 | 47703802 | 49098329 | 70675229 | 70187705 | 94810607 | 93494872 | 241758253 | 232116347 | 719102035 | 708609392 | 1171749070 |
| | | SAM-0.2 | 47939222 | 48133948 | 72101733 | 71230971 | 96604004 | 94827507 | 239904644 | 237095046 | 717639244 | 716205389 | 1191606245 |

The performance of dynamic dataset group results is visually compared in Figure 15. For small benchmark dataset group instances GK01 and GK02, the Full-Restart strategy performs the best with an average result gap of 0.14% and 0.18%, respectively. This strategy solves each optimization state from the start, and information learned from previous states does not negatively impact algorithm convergence. The ACO with Aphids strategy demonstrated superior performance in comparison to Full-Restart and Pheromone-Sharing strategies for all larger instances GK03 through GK11. Additionally, the ACO with Aphids strategy outperforms the Pheromone-Sharing strategy for every dynamic dataset group.

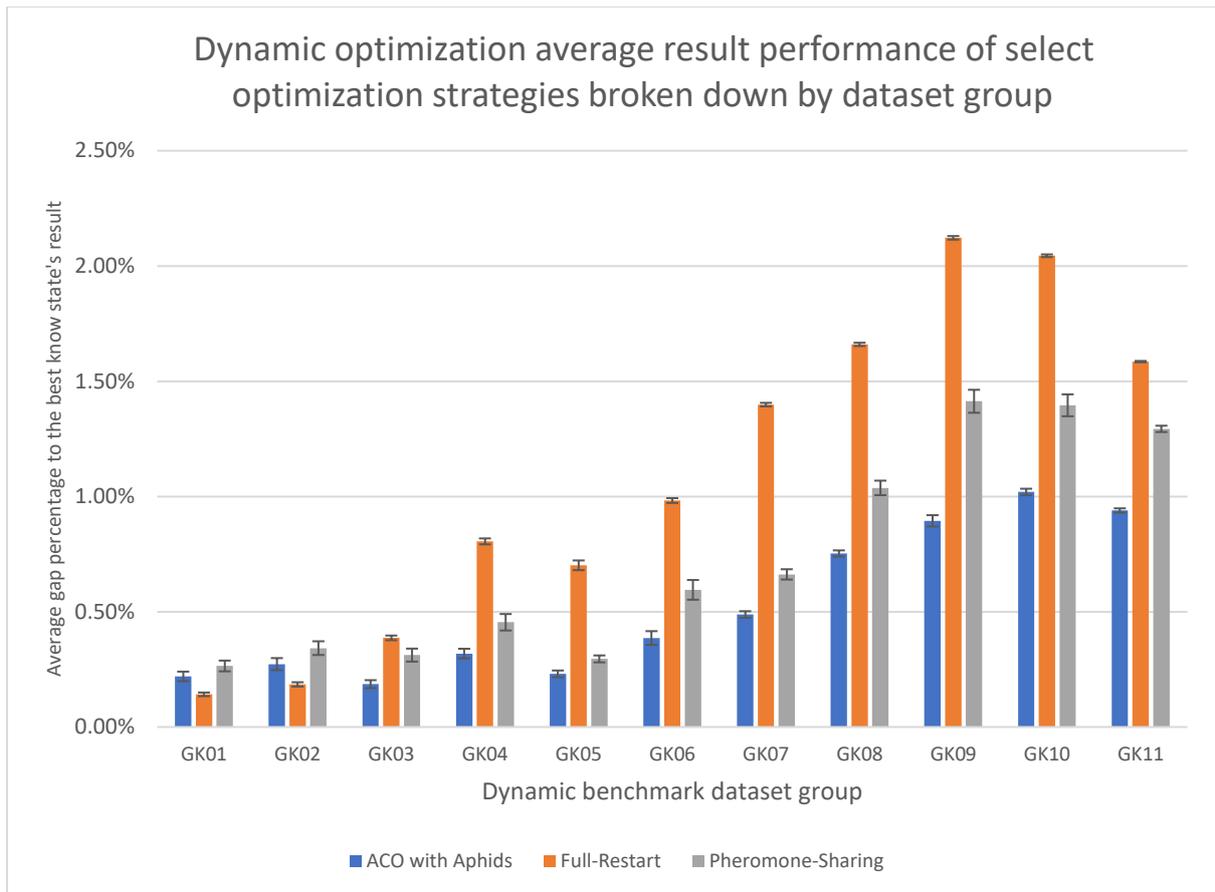

*Figure 15: Dynamic optimization average performance of each dynamic optimization strategy averaged per dataset group. Each data point is an average result gap of all five dynamism levels run ten times. Error bars indicate the standard deviation of experiment results.*

The average performance of the state's first and the last iterations across all dynamic dataset groups are compared in Figure 16. The upper mark indicates the average result gap achieved within the first iteration after the dynamic change, and the lower mark is the average final result achieved before the dynamic change occurs. The first iteration of the Full-Restart strategy for all dataset groups has the worst result gap of 2.51%. This behaviour is expected because the Full-Restart strategy represents the worst-case scenario where the information is not carried from one dynamic state to the next, and each state has to converge independently. However, the Full-Restart strategy also improves the result gap the most by an average of 1.42%. The first iteration result gap of the ACO with Aphids strategy is, on average, 0.83%, which is only slightly better than the result gap of the Pheromone-Sharing strategy of 0.99%. However, the improvement of the ACO with Aphids strategy from the first iteration to the last by 0.31% is more significant than the improvement of Pheromone-Sharing strategy by 0.25% for every dataset group. This indicates that aphids help ants adapt to a new dynamic environment quicker, and the starting pheromone is less localized to an outdated solution. The benefit of using aphids is compounded for especially large benchmark dataset groups, like GK08 and larger.

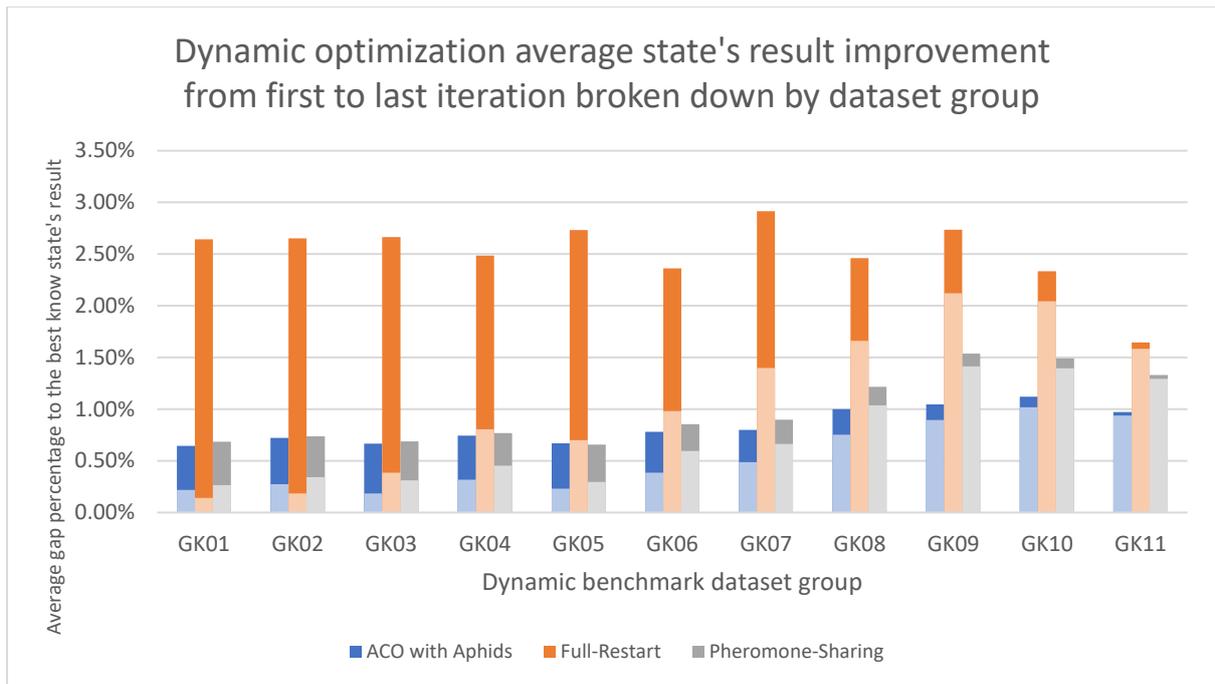

*Figure 16: Dynamic optimization average state's result improvement from the first to the last iteration of select optimization strategies broken down by dataset group. Each result data point is an average result gap of all five dynamism levels run ten times.*

Selected optimization strategies are also compared by performance for all dynamism levels in Figure 17. The Full-Restart strategy performs almost equally well for all dynamism levels, with an average result gap of 1.09%. This behaviour is expected as the Full-Restart strategy solves each dynamic state independently, and dynamism level has no impact on algorithm performance. Then for both ACO with Aphids and Pheromone-Sharing strategies, a lower dynamism level allows for better performance because previously found solutions are changed to a lower degree and are more up-to-date. On the flip side, the larger dynamism level hurts the dynamic optimization performance of ACO with Aphids and Pheromone-Sharing strategies. At the highest dynamism level, $\Delta = 0.2$ Pheromone-Sharing strategy is no longer beneficial over the Full-Restart strategy. Furthermore, the ACO with Aphids strategy consistently outperforms the Pheromone-Sharing strategy across all dynamism levels, on average 29.2% lower result gap, the lowest reduction for SAM-0.01 dynamism level is 24.4%, and the highest reduction for SAM-0.1 dynamism level is 31.4%.

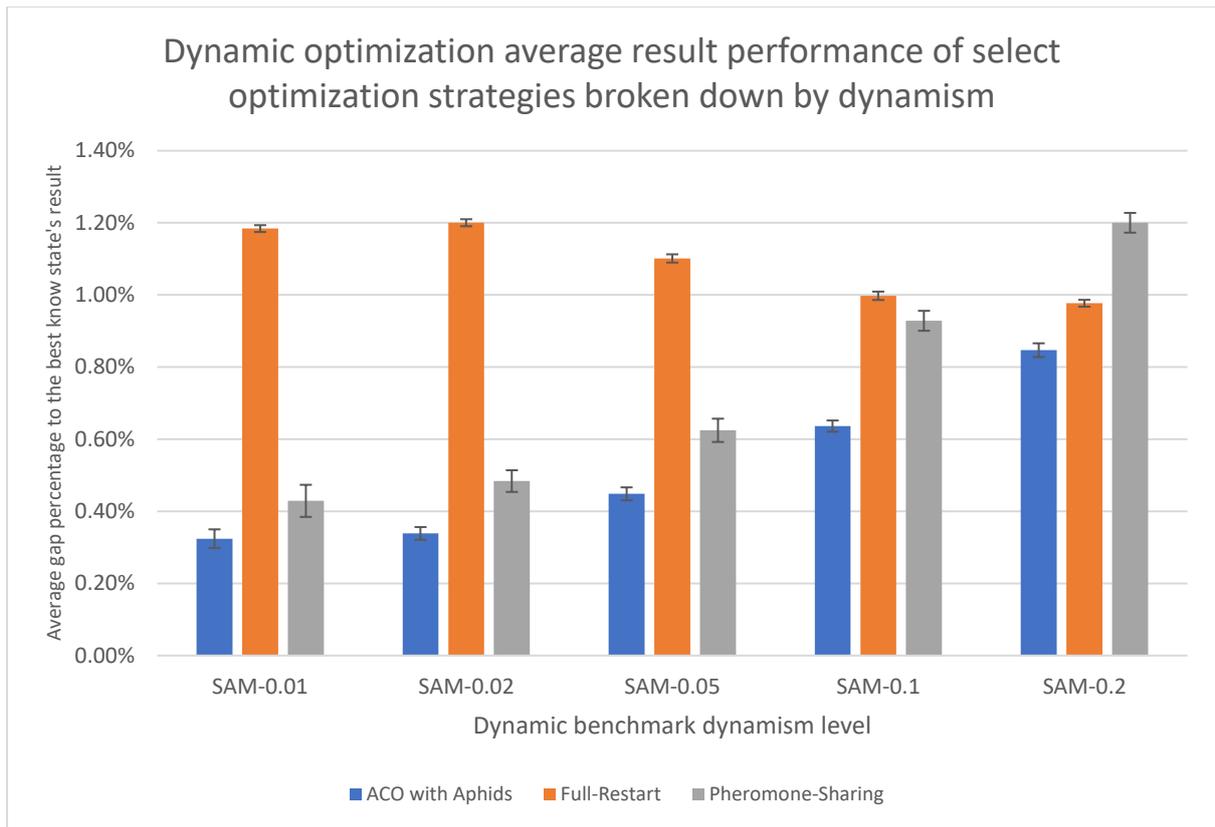

*Figure 17: Dynamic optimization average performance of each dynamic optimization strategy averaged per dynamism. Each data point is an average result gap of all 11 dataset groups run ten times. Error bars indicate the standard deviation of experiment results.*

The average performance of the first and last iteration of the state across all dynamism levels is compared in Figure 18. Similarly to Figure 16, the upper mark indicates the average result gap achieved within the first iteration after the dynamic change has occurred, and the lower mark is the average final result achieved before the dynamic change occurs. As expected for the Full-Restart strategy, dynamism has almost no impact on the first iteration's performance after the dynamic change because each state is solved independently. ACO with Aphids and Pheromone-Sharing strategies, higher dynamism causes more prominent performance degradation after the dynamic change. However, on average, the overall first iteration performance of the ACO with Aphids strategy is better by 15.6% than the Pheromone-Sharing strategy.

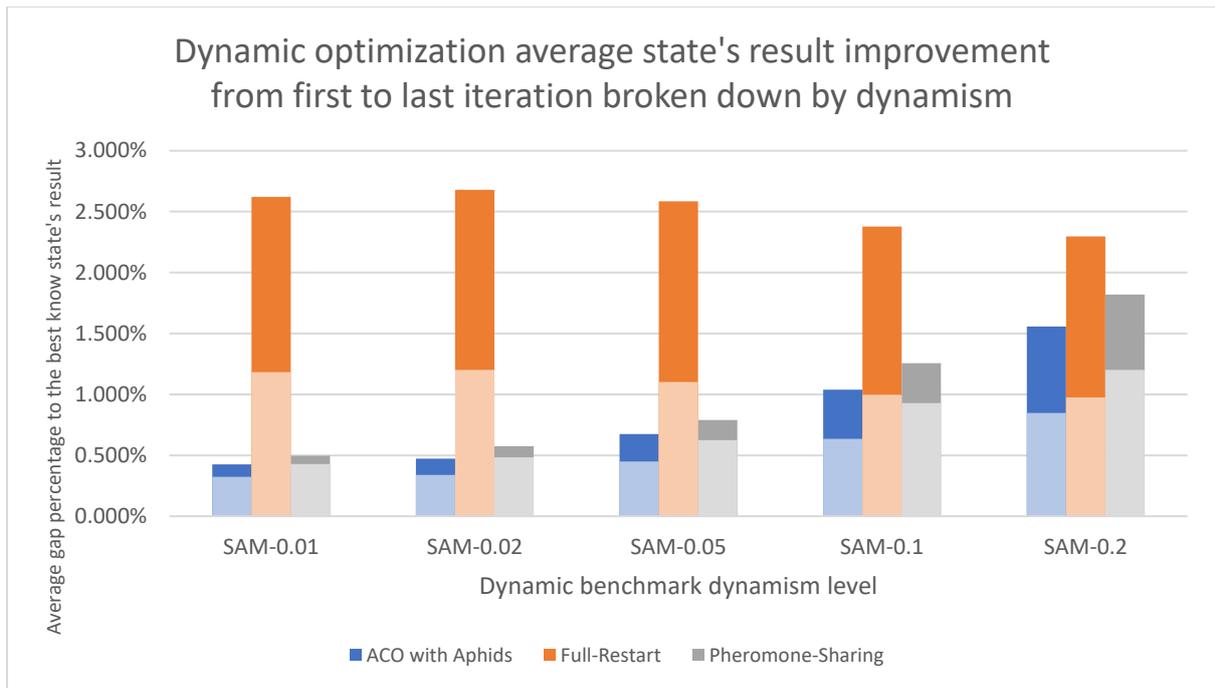

*Figure 18: Dynamic optimization average state's result improvement from the first to the last iteration of select optimization strategies broken down by dynamism. Each result data point is an average result gap of all 11 dataset groups run ten times.*

Finally, the aggregate convergence through all states is compared in Figure 19. Each convergence line represents the average convergence of all tested benchmark datasets. The average convergence is calculated from the optimization result with a normalized optimization time to account for differences in the time given for each state's optimization based on benchmark dataset complexity. The Pheromone-Sharing strategy and ACO with Aphids strategy reach equilibrium over the first few states where subsequent state solutions results are no closer to best-know solution results than previous states' solutions. Meanwhile, the Full-Restart strategy does not show any inter-state convergence.

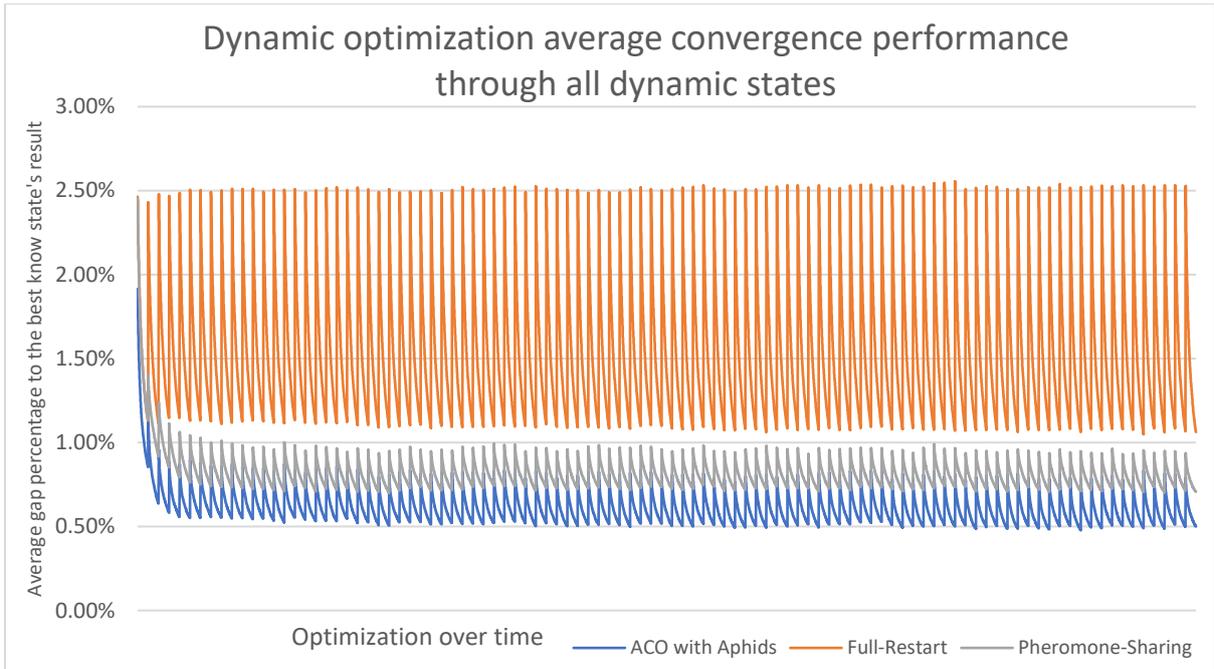

*Figure 19: Dynamic optimization average convergence performance through all dynamic states. Each convergence performance result is an average convergence of all 55 benchmark datasets run ten times.*

For a clearer view, Figure 20 shows the same aggregate convergence as in Figure 19, but only through the first ten states. The ACO with Aphids strategy starts with significantly better first state optimization results than Full-Restart and Pheromone-Sharing strategies. This first state's convergence improvement is caused by a high aphids' relocation rate, which occurs in every state, including the first. Then as predicted in Figure 7, the Full-Restart strategy converges almost equally for every state. The Pheromone-Sharing strategy has a minimal result gap slip and a reduced convergence slope. Finally, ACO with Aphids strategy has reasonably good result gap slip while maintaining an excellent convergence slope after each dynamic change.

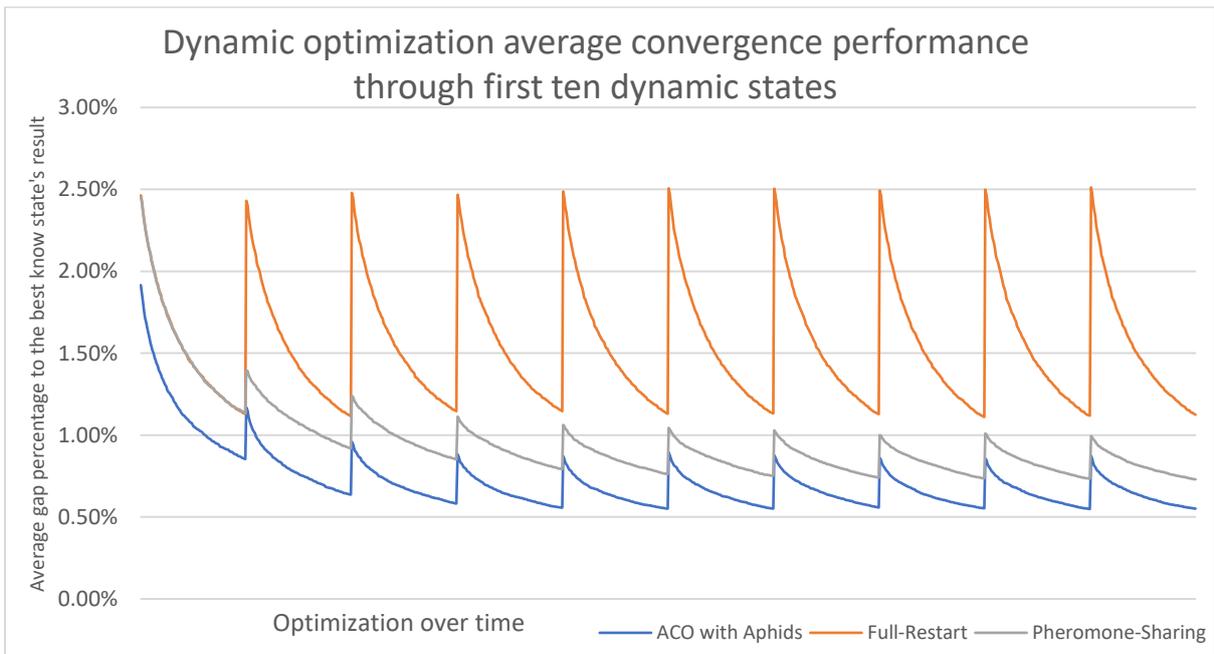

*Figure 20: Dynamic optimization average convergence performance through the first ten dynamic states. Each convergence performance result is an average convergence of all 55 benchmark datasets run ten times.*

In summary, ACO with Aphids has outperformed the Pheromone-Sharing strategy and has proven especially beneficial for large optimization problems. This demonstrates how adding aphids to the ACO algorithm improves the dynamic performance of large optimization problems with limited time allowed to converge. Also, for the smallest optimization problems Full-Restart strategy has performed better than dedicated strategies for dynamic optimization, as it was accurately predicted by previous research [27]. ACO with Aphids strategy is able to achieve improved solution quality Pheromone-Sharing strategy primarily due to decoupled pheromone that ants use from the information shared among the dynamic states. The Pheromone-Sharing strategy reuses the pheromone matrix left from the previous state, while Aphids of the ACO with Aphids create a new pheromone matrix better suited for the new dynamic optimization state. Aphids are able to create better pheromones utilizing a combination of two factors: an update based on the previous best dynamic state results and an update based on heuristic information of the new dynamic state.

# 6   Conclusions and future directions

This paper aimed to solve the trade-off problem offered by current ACO dynamic optimization strategies, where the Full-Restart strategy shows a significant penalty to solution quality after each dynamic change, and the Pheromone-Sharing strategy has a reduced slope of convergence.

This research has proposed nature-inspired addition to the Ant Colony Optimization algorithm to improve its performance for discrete dynamic optimization problems. Proposed method modelled ants' interaction with aphids in the dynamic environment. In the real world, Aphids produce honeydew which is nutritious to ants, and under ants' influence, aphids give up their mobility to increase honeydew production. This nature-inspired interaction between ants and aphids is beneficial for dynamic optimization, where ants control the population of the aphids and placed aphids mediate the information sharing across dynamic states of the optimization problem.

Then ACO with Aphids algorithm has been tested against the two most popular dynamic optimization strategies, Full-Restart and Pheromone-Sharing on Dynamic Multidimensional Knapsack Problem (DMKP). ACO with Aphids has significantly outperformed the Full-Restart strategy for large dataset groups and a limited amount of time to solve each state. On average, the result gap was reduced by 52.5%, which is a 110% better performance. Also, ACO with Aphids has outperformed the Pheromone-Sharing strategy in every optimization scenario for all dynamism levels and dataset group sizes. On average, the result gap was reduced by 29.2%, which is a 41% better performance. The test results have proved ACO with Aphids superior performance over both Full-Restart and Pheromone-Sharing strategies with rejected null hypothesis, P-value less than $1^{-6}$.

Overall, the proposed ACO with Aphids algorithm proved to be a well-rounded, dynamic optimization strategy with a strong ability to adapt to dynamic change and maintain quick convergence. This strong adaptability further compounds through several dynamic states for especially large optimization problems, where positive convergence occurs over multiple dynamic optimization states.

ACO with Aphids algorithm has only been compared to the other two ACO dynamic optimization strategies. In the future, researchers aim to perform dynamic optimization tests to compare ACO with Aphids algorithm to other well-known dynamic optimization algorithms based on GA and PSO. Another important research direction is comparing test ACO with Aphids performance to solve real-world dynamic optimization problems.